\documentclass{article}

\usepackage[final]{corl_2022} 

\usepackage{graphicx}
\usepackage{svg}
\usepackage{subcaption}
\usepackage{alphalph}

\usepackage{amsmath} 
\usepackage{adjustbox}
\usepackage{amsthm}

\usepackage{hyperref}
\usepackage{amssymb}  
\usepackage{algorithm}
\usepackage[noend]{algpseudocode}
\algnewcommand\algorithmicinput{\textbf{Input:}}
\algnewcommand\INPUT{\item[\algorithmicinput]}
\algnewcommand\algorithmicoutput{\textbf{Output:}}
\algnewcommand\OUTPUT{\item[\algorithmicoutput]}
\usepackage{subfiles}

\usepackage{enumitem}
\usepackage{color}

\usepackage[font=footnotesize,belowskip=-10pt,aboveskip=3pt]{caption}
\usepackage[capitalise]{cleveref}

\newtheorem{prop}{Proposition}[section]
\newtheorem{definition}{Definition}[section]

\DeclareMathOperator*{\minimize}{minimize}
\DeclareUnicodeCharacter{2212}{-}

\usepackage{todonotes}

\newcommand{\modelname}[0]{Risk-Aware Prediction}
\newcommand{\modelacronym}[0]{RAP}

\title{\modelacronym: \modelname~for Robust Planning}

\author{
  Haruki Nishimura$^{*}$
  \And
  Jean Mercat\thanks{The first two authors contributed equally to this work.}
  \And
  Blake Wulfe
  \And
  Rowan McAllister
  \And
  Adrien Gaidon \\
  Toyota Research Institute, USA \\
  \texttt{firstname.lastname@tri.global}
}

\begin{document}
\maketitle


\begin{abstract}
Robust planning in interactive scenarios requires predicting the uncertain future to make risk-aware decisions.
Unfortunately, due to long-tail safety-critical events, the risk is often under-estimated by finite-sampling approximations of probabilistic motion forecasts.
This can lead to overconfident and unsafe robot behavior, even with robust planners.
Instead of assuming full prediction coverage that robust planners require, we propose to make prediction itself risk-aware.
We introduce a new prediction objective to learn a risk-biased distribution over trajectories, so that risk evaluation simplifies to an expected cost estimation under this biased distribution.
This reduces the sample complexity of the risk estimation during online planning, which is needed for safe real-time performance.
Evaluation results in a didactic simulation environment and on a real-world dataset demonstrate the effectiveness of our approach. The code\footnote{\url{https://github.com/TRI-ML/RAP}} and a demo\footnote{\url{https://huggingface.co/spaces/TRI-ML/risk_biased_prediction}} are available.
\end{abstract}


\section{Introduction}
\label{sec: introduction}
In safety-critical and interactive control tasks such as autonomous driving, the robot must successfully account for uncertainty of the future motion of surrounding humans.
To achieve this, many contemporary approaches decompose the decision-making pipeline into prediction and planning modules~\cite{schmerling2018multimodal, fan2018baidu, zeng2019end, ivanovic2020mats, cui2021lookout} for maintainability, debuggability, and interpretability. A prediction module, often learned from data, first produces likely future trajectories of surrounding agents, which are then consumed by a planning module for computing safe robot actions.
Recent works~\cite{nishimura2020risk, novin2021risk} further propose to couple prediction with risk-sensitive planning for enhanced safety, wherein the planner computes and minimizes a risk measure~\cite{shapiro2021lectures} of its planned trajectory based on probabilistic forecasts of human motion from the data-driven predictor.
A risk measure is a functional that maps a cost distribution to a deterministic real number, which lies between the expected cost and the worst-case cost~\cite{majumdar2020should}.

Although combining data-driven forecasting and risk-sensitive planning has been shown to be effective, there exist several limitations to this approach.
First, accurate risk evaluation of candidate robot plans remains challenging, due to inaccurate characterization of uncertainty in human behavior~\cite{cheng2021limits} and finite-sampling from the predictor.
Some existing methods that promote diversity of prediction (e.g.,~\cite{yuan2019diverse, huang2020diversitygan}) may alleviate this issue, but they are not explicitly designed for reliable risk estimation needed for robust planning.
Second, endowing an existing planner with risk-sensitivity often requires non-trivial modifications to its internal optimization algorithm~\cite{shaiju2008formulas, bauerle2011markov, chow2015risk}.
This modification can be problematic, if, for example, an autonomy stack already has a dedicated and complex (risk-neutral) planner in use and cannot easily modify its internal optimization algorithms.

To address the above limitations, we propose to consider risk within the \textit{predictor} rather than in the planner.
We present a risk-biased trajectory forecasting framework, which provides a general approach to making a generative trajectory forecasting model risk-aware. 
Our novel method augments a pre-trained generative model with an additional encoding process.
This modification changes the output of the prediction so that it purposefully and deliberately over-estimates the probability of dangerous trajectories.
This ``pessimistic" forecasting model gives \emph{distributional robustness} (e.g.,~\cite{rahimian2019distributionally}) to the planner against potential inaccuracies of the human behavior model.

We achieve the pessimistic risk-biased distribution using a novel prediction loss. This shifts the computational burden of drawing many prediction samples that capture rare events from online deployment to offline prediction training.
The planner can still obtain an accurate estimate of the risk measure in real-time during deployment with fewer prediction samples required from the biased distribution.
Furthermore, our approach also eliminates the need for modifications to the planner’s optimization algorithm.
Thus, one can achieve enhanced safety by simply replacing a conventional probabilistic motion forecaster with the proposed risk-biased model, while still using the same existing risk-neutral planner.
This capability is intended for use in robotic applications where misestimation of risk could lead to injury, including autonomous vehicles and home robots that must operate safely in close proximity to humans.

Specifically, our contributions in this work are as follows:
\begin{itemize}
    \item We propose a risk-biased trajectory forecasting framework, which makes forecasts more useful for the downstream task and leads to plans that are robust to distribution shifts.
    \item Our risk-biased model off-loads the heavy computation of risk estimation from online planning, providing risk-awareness to a generic risk-neutral planner.
    \item We extensively evaluate our proposed approach in simulation with a planner in the loop and offline with complex real-world data. 
\end{itemize}

%

\section{Related Work}
\label{sec: related_work}
\paragraph{Trajectory forecasting from data.}

Early trajectory forecasting approaches defined hand-crafted dynamics models~\cite{mercat2019kinematic, scholler2020constant}, and incorporated rules that induce obstacle avoidance behavior~\cite{helbing1995social} or mimic the overall traffic flow~\cite{treiber2000congested, kesting2007general}. More recently, data-driven, learning-based methods have gained popularity for their ability to better capture the complexity of human behavior~\cite{lefevre2014survey}, and typically use neural networks defining multi-modal trajectory distributions~\cite{gupta2018social, deo2018multi, bhattacharyya2018accurate,  huang2020diversitygan, ngiam2021scene,huang2021tip,  salzmann2020trajectron++, gilles2021gohome, gilles2021thomas, sadeghian2019sophie, mercat2020multi,  rhinehart2018r2p2,  gilles2021home,amirian2019social, zhao2020tnt, narayanan2021divide, varadarajan2021multipath++}.

Significant effort is directed toward increasing the coverage, or diversity, of motion forecasting models \cite{lee2017desire, rhinehart2018r2p2, amirian2019social, gilles2021home, casas2020implicit, zhao2020tnt, varadarajan2021multipath++, narayanan2021divide, yuan2019diverse, yuan2020dlow, huang2020diversitygan} in order to ensure that no critical events are missed.
Diversity can be explicitly encouraged using a best-of-many loss~\cite{bhattacharyya2018accurate}, by replacing a mean-squared loss with a Huber loss~\cite{casas2020implicit}, by choosing trajectory samples that maximize the distribution coverage~\cite{gilles2021home}, or by setting diverse anchors or target points~\cite{zhao2020tnt, varadarajan2021multipath++, narayanan2021divide}.
Another strategy to increase mode coverage takes advantage of the latent distribution of CVAEs~\cite{yuan2019diverse, yuan2020dlow, cui2021lookout} or GANs~\cite{huang2020diversitygan}.
\citet{cui2021lookout} argue that besides coverage, sample efficiency is also an important factor. The authors trained a road-scene motion forecasting model to produce predictions of other agents that induce diverse reactions from the given robot planner.
Similarly, \citet{mcallister2022control} train a model with a weighted loss giving a low weight to the predictions that do not affect the planner.
\citet{huang2021tip} train a forecasting model that allows a simple optimization procedure to select the safest among a set of plans generated by a planner.
While prior work considered task-awareness or planner-awareness, to the best of our knowledge, we are the first to use risk as a proxy to make forecasts more useful for the downstream task.

\paragraph{Subjective probability and prospect theory.}
Our pessimistic risk-biased prediction can be interpreted as a model of subjective probability (e.g.,~\cite{savage1972foundations}), which is closely related to risk-awareness~\cite{shen2014risk}.
For instance, prospect theory~\cite{tversky1992advances} studies how humans make risk-aware decisions and introduces the notion of \emph{probability weighting}~\cite{barberis2013thirty}.
Under this model, the distribution is ``warped" so that the probabilities of unlikely events are always over-weighted.
Recent robotics literature has leveraged prospect theory to better model risk-awareness in human decision making, for example, in collaborative human-robot manipulation~\cite{kwon2020humans} and driver behavior modeling~\cite{sun2019interpretable}.

Prospect theory is a descriptive model of human decision making, which differs from our goal of designing risk-aware robots.
Moreover, our model only overestimates the probability of events that incur high-cost for the robot, unlike probability weighting that overestimates any unlikely outcome.

\paragraph{Risk-sensitive planning and control.}
Risk-sensitive planning and control date back to the 1970s, as exemplified by risk-sensitive Linear-Exponential-Quadratic-Gaussian~\cite{jacobson1973optimal, whittle1981risk} and risk-sensitive Markov Decision Processes (MDPs)~\cite{howard1972risk}.
More recent methods include risk-sensitive nonlinear MPC~\cite{roulet2020convergence, nishimura2020risk}, Q-learning~\cite{borkar2002q, shen2014risk}, and actor-critic~\cite{chow2014algorithms, choi2021risk} methods, for various types of risk measures.
Refer to a recent survey~\cite{wang2021risk} for further details.
Unlike those methods in which the policy directly optimizes a risk-measure, we propose to instead bias the prediction so that risk-sensitivity can be achieved by a risk-neutral planner that simply optimizes the expected value of the cost.

\section{Background}
\label{sec: background}
\subsection{Generative Probabilistic Trajectory Forecasting}
\label{subsec: generative_forecasting}
Let $x$ and $y$ be the past and the future trajectories of an agent, and $Y|_x$ denote the random variable of the future trajectory conditioned on the observed past trajectory $x$. 
We would like to fit the distribution of $p(Y|_x)$ given a dataset $\mathcal{D}$ of i.i.d.\ samples of $(x, y)$ pairs. To fit $p(Y|_x)$, we maximize the likelihood of future trajectories w.r.t.\ the model parameters $\theta, \phi$:
\smash{${\operatorname{maximize}}_{\theta, \phi}~  \prod_{(x, y) \in \mathcal{D}}\mathcal{L}(\theta, \phi ;y|_x)$},
where $\mathcal{L}(\theta, \phi ;y|_x)$ is the likelihood of the sample $y$ knowing $x$.
One method to fit this distribution is to learn a conditional variational auto-encoder, CVAE \cite{sohn2015learning}.
We focus on this approach because it produces a structured latent representation.
The CVAE conditions its likelihood estimation on a latent random variable $Z|_{x, y}$ with a posterior $q_{\phi_2}(z|_{x,y})$, or $Z|_x$ with an inferred prior $q_{\phi_1}(z|_x)$ used in the joint likelihood $p_{\theta}(y|_x, z|_x)$.
The marginal likelihood of the future trajectory (or ``model evidence'') is $p_{\theta}(y|_{x, z})$, and can be rewritten as: 
\begin{equation}
    \begin{aligned}
         \mathcal{L}(\theta, \phi ;y|_x) 
         \;=\; \int p_{\theta}(y|_{x , z}) dz 
         \,=\, \int p_{\theta}(y|_{x, z}) \frac{q_{\phi_2}(z|_{x,y})}{q_{\phi_2}(z|_{x,y})} dz
         \,=\, \mathbb{E}_{q_{\phi_2}(z|_{x,y})} \left[ \frac{p_{\theta}(y|_{x}, z|_{x})}{q_{\phi_2}(z|_{x,y})} \right].
    \end{aligned}
    \label{eq: likelihood_latent}
\end{equation}
Using Jensen's inequality, the logarithm of \eqref{eq: likelihood_latent} is lower bounded by
\begin{equation}
        \begin{aligned}
             L(\theta, \phi ; x, y)
            \;=\; \mathbb{E}_{q_{\phi_2}(z|_{x,y})} \left[ \ln( p_{\theta}(y|_{x, z}) ) \right] \,-\, \mathrm{KL} \big(q_{\phi_2}(z|_{x,y}) || q_{\phi_1}(z|_x) \big),
        \end{aligned}
    \label{eq: lower_bound}
\end{equation}
called the evidence lower bound (ELBO).
We model $q_\phi$ and $p_\theta$ using neural networks. The encoders assume a Gaussian prior with independent elements to produce the inferred prior $f_{\phi_1}:(x) \to (\mu|_x, \operatorname{diag}(\Sigma|_x))$, and the posterior $f_{\phi_2}:(x, y) \to (\mu_{x,y}, \operatorname{diag}(\Sigma|_{x,y}))$. The decoder makes the forecast $g_\theta: (x, z) \to y$.
Every term in \eqref{eq: lower_bound} can be either computed or estimated with Monte-Carlo sampling as established in \cite{kingma2013auto, sohn2015learning}.

\subsection{Risk Measures}
\label{subsec: risk_measures}
A risk measure is defined as a functional that maps a cost distribution to a real number. 
In other words, given a random cost variable $C$ with distribution $p$, a risk measure of $p$ yields a deterministic number $r$ called the risk.
In practice, we often consider a class of risk measures that lie between the expected value $\mathbb{E}_p[C]$ and the highest value $\sup(C)$.
The former corresponds to the risk-neutral evaluation of $C$, while the latter gives the worst-case assessment.
Such risk measures often take a user-specified risk-sensitivity level $\sigma \in \mathbb{R}$ as an additional argument, which determines where the risk value $r$ is positioned between $\mathbb{E}_p[C]$ and $\sup(C)$.
Formally, let us define a risk measure as $\mathcal{R}_p: (C, \sigma) \to r \in [\mathbb{E}_p[C], \sup(C)]$.
Examples of such risk measures include entropic risk~\cite{whittle1981risk}: $\mathcal{R}_p^{\text{entropic}}(C, \sigma) = \frac{1}{\sigma}\log\mathbb{E}_p[\exp(\sigma C)]$
as well as CVaR~\cite{pflug2000some}:
\begin{align}
    \mathcal{R}_p^{\text{CVaR}}(C, \sigma) \;=\; \inf_{t \in \mathbb{R}} \left\{t + \frac{1}{1 - \sigma} \mathbb{E}_p\left[\max(0, C - t)\right]\right\}.
    \label{eq: cvar}
\end{align}
The rest of the paper assumes CVaR \eqref{eq: cvar} as the underlying risk measure, but note that the proposed approach is not necessarily bound to this particular choice.
For CVaR, the risk value $r$ given risk-sensitivity level $\sigma \in (0, 1)$ can be interpreted as the expected value of the right $(1 - \sigma)$-tail of the cost distribution \cite{trindade2007financial}. Thus, $\mathcal{R}_p(C, \sigma)$ tends to $\mathbb{E}_p[C]$ as $\sigma \to 0$ and to $\sup(C)$ as $\sigma \to 1$.

Another intriguing property of CVaR is its fundamental relation to distributional robustness.
CVaR belongs to a class of risk measures called \emph{coherent measures of risk}~\cite{rockafellar2007coherent} with the following dual characterization (\cite{rockafellar2007coherent}, Theorem 4a):
\begin{align}
    \mathcal{R}_p(C, \sigma) \;=\; \sup_{q \in \mathcal{Q}} \mathbb{E}_q[C],
    \label{eq: distributional_robustness}
\end{align}
where $\mathcal{Q}$ is a uniquely-determined, non-empty and closed convex subset of the set of all density functions.
This suggests that CVaR is equivalent to a worst-case expectation of the cost $C$ when the underlying probability distribution $q$ is chosen adversarially from $\mathcal{Q}$.
Therefore, an autonomous robot optimizing CVaR (or coherent measures of risk in general) obtains distributional robustness, in that the objective accounts for robustness to potential inaccuracies in the underlying probabilistic model.
In this context, the set $\mathcal{Q}$ is often referred to as an \emph{ambiguity set} in the literature~\cite{van2015distributionally, samuelson2017data}.

\section{Problem Formulation}
\label{sec: problem_formulation}
Suppose that a robot incurs cost $C$ under a planned policy $\pi$ or trajectory.
This cost is given by a function $J^{\pi}$ such that $C = J^{\pi}(Y)$ with $Y$ being the human future trajectory random variable, which the robot predicts probabilistically.
We assume that $J^{\pi}$ is known and differentiable in $y$ for each $\pi$.
One can design such a cost function so that $J^{\pi}(y)$ is high when the robot collides into the particular trajectory $Y = y$ of a human. Supplementary material \ref{sec: ttc_cost} defines the cost function used in this work.

We begin with a pre-trained generative model, as defined in Section \ref{subsec: generative_forecasting}, that gives a predictive distribution $p(Y|_x) = \int p(Y |_{x, z}) p(z)dz$ through an inferred latent distribution $p(Z|_x)$. This latent is mapped to the trajectory space by a generator or decoder $y = g(z, x)$.
Under this unbiased model, the risk is given by $r = \mathcal{R}_p\left(J^{\pi}(g(Z, x)), \sigma\right)$ using the risk measure introduced in Section~\ref{subsec: risk_measures}.

Given the unbiased model and the risk measure, we are interested in finding another distribution $q_{\psi}(Z)$ in the latent space with learnable parameters $\psi$, under which simply taking the risk-neutral expectation of the cost will yield the same risk value as given above. This can be achieved by enforcing the following equality constraint on this \emph{biased} distribution $q_{\psi}(Z)$:
\begin{equation}
    \mathbb{E}_{q_\psi}\left[J^{\pi}(g(Z, x))\right] \;=\; \mathcal{R}_p\left(J^{\pi}(g(Z, x)), \sigma\right).
    \label{eq:risk_constraint}
\end{equation}
We show that such a distribution exists in Section \ref{prop: existence} of the supplementary material.
Comparing both sides in \eqref{eq:risk_constraint}, we note that such $q$ should be dependent on the risk-sensitivity level $\sigma$.
We propose to optimize the parameters $\psi$ of the risk-biased distribution $q_\psi(Z|_{x, \sigma})$.
In general, many distributions $q$ can satisfy \eqref{eq:risk_constraint}.
We propose to pick a particular $q$ that additionally minimizes the KL divergence from the prior $p$, to prevent the biased distribution from becoming too different from the original unbiased distribution.
This leads to the following constrained optimization problem:
\begin{equation}
    \underset{\psi}{\operatorname{\minimize}} \quad \mathrm{KL}\left(q_{\psi}(Z|_{\sigma}) \Vert p(Z)\right) \quad
    \textrm{subject to} \; \mathbb{E}_{q_\psi}\left[J^{\pi}(g(Z, x))\right] \,=\, \mathcal{R}_p\left(J^{\pi}(g(Z, x)), \sigma\right).
    \label{eq:optimization_problem}
\end{equation}
In general, we cannot guarantee uniqueness of the solution to the optimization problem \eqref{eq:optimization_problem}. However, in the supplementary material \ref{sec: proofs}, we provide further analysis of \eqref{eq:optimization_problem} along with a sufficient assumption under which the solution would be unique (Proposition \ref{prop: unicity_vp}).

\paragraph{Connection to importance sampling.} Importance sampling has been employed in rare-event simulation for accelerated safety verification of autonomous systems~\cite{zhao2017accelerated, wulfe2018real, o2018scalable}, which yields
a pessimistic sampling distribution similar to our risk-biased model.
However, a crucial difference of our approach is that it estimates a more general risk measure instead of an expected value.
Given a desired risk-sensitivity level, unweighted samples from the proposal $q$ will directly yield the risk estimate \eqref{eq:risk_constraint}.
This removes the need to compute the importance weights.

\paragraph{Connection to distributional robustness.}
When a coherent measure of risk is chosen as the underlying risk measure (such as CVaR), the right-hand side of \eqref{eq:risk_constraint} is always equivalent to a worst-case distribution $q$ chosen out of an ambiguity set $\mathcal{Q}$ \eqref{eq: distributional_robustness}. In general, it is difficult to verify if the optimal distribution $q_{\psi^*}$ is in $\mathcal{Q}$, since the specifics of $\mathcal{Q}$ depend on the choice of the risk measure as well as the risk-sensitivity level $\sigma$. Nevertheless, it holds true that any feasible distribution $q_{\psi}$ for \eqref{eq:optimization_problem} yields the same worst-case expected cost as the most adversarial distribution from $\mathcal{Q}$.
Therefore, a planner relying on $q_{\psi}$ instead of $p$ will possess distributional robustness.
We demonstrate this crucial capability via an empirical evaluation in Section \ref{sec:distributional_robustness_test}.

\section{Implementation Details}
\label{sec: implementation}

Section \ref{sec: architecture} of the supplemental defines a usual (unbiased) CVAE trajectory forecasting model that learns two encoders, defining the Gaussian latent variables $Z|_x$ and $Z|_{x,y}$, and one decoder, predicting $Y|_{x, z}$.
We propose to solve problem \eqref{eq:optimization_problem} by learning a third neural network encoder to define a biased latent distribution that, in combination with the pre-trained decoder, produces biased forecasts.
This biased encoder takes the past trajectory $x$, a risk-level $\sigma$, and the robot future trajectory $y_\text{robot}$. It outputs the parameters of a Normal distribution $\mu^{(b)}$ and $\log(\operatorname{diag}(\Sigma^{(b)}))$.

In practice, we soften the hard constraint \eqref{eq:risk_constraint} by using the penalty method~\cite{kochenderfer2019algorithms}, which progressively increases the weight $\alpha$ of the risk-loss during training.
We also leverage a user-defined sampling distribution $p(\sigma)$ to sample different risk-sensitivity levels during training, so that the risk estimate remains accurate at any reasonable value of $\sigma$ at inference time.
Finally, we encourage the model to overestimate the risk rather than underestimate it so we scale by the positive value $s$ and define an asymmetric risk-loss that penalizes linearly the underestimation of the risk and logarithmically its overestimation:
\begin{equation}
    \rho(x) \;=\; 
    \begin{cases}
    s|x|,& \text{if } s x\leq 1\\
    \log(s x),  & \text{otherwise.}
\end{cases}
\end{equation}
We obtain the following loss function with $\alpha$ and $\beta$ controlling the relative importance of the losses:
\begin{align*}
    \mathcal{L}(\psi) \;=\; \mathbb{E}_{\sigma \sim p(\sigma)}\left[  \alpha \,\rho\big(\mathbb{E}_{q_\psi}\left[J^{\pi}(g(Z, x))\right] - \mathcal{R}_p\left(J^{\pi}(g(Z, x)), \sigma\right)\big) + \beta\, \mathrm{KL}\left(q_{\psi}(Z|_{\sigma, x}) \Vert p(Z|_x)\right)\right].
\end{align*}
The expected values and the risk measure are approximated by Monte Carlo sampling.
For computing CVaR ($\mathcal{R}_p\left(J^{\pi}(g(Z, x)), \sigma\right)$ ), we use the estimator proposed by Hong et al.~\cite{hong2014monte}. Consistency and asymptotic normality of this estimator hold under mild assumptions~\cite{hong2014monte}.

{\setlength{\textfloatsep}{5pt}
\begin{algorithm}[t]
    \footnotesize
    \caption{Proposed Risk-Biasing Loss Estimation}
    \label{algo: train_proc_1}
    \begin{algorithmic}[1]
        \INPUT Trajectory $(x, y) \sim \mathcal{D}$, risk level $\sigma \sim p(\sigma)$, KL-loss weight $\beta$, risk weight $\alpha$, robot motion $y_{\text{robot}}$
        \For{$k \in \{1,\dots,K_1\}$}
            \State Sample latent $z_k|_x \sim \mathcal{N}(\mu|_x, \Sigma|_x)$ with prior parameters ($\mu|_x$, $\Sigma|_x) = f_{\phi_1}(x)$
            \State Decode risk-neutral predictions $y_k = g_\theta(x, z_k|_x)$
        \EndFor
        \State Compute risk $r$ using $\{y_1,\dots y_{K_1}\}$ and $J^{y_{\text{robot}}}$ with Monte Carlo estimation (e.g.,~\cite{hong2014monte})
        \For{$k \in \{1,\dots,K_2\}$}
            \State Sample biased latent $\hat{z}_k^{(b)} \sim \mathcal{N}(\mu^{(b)}, \Sigma^{(b)})$ with risk-biased parameters $(\mu^{(b)}, \Sigma^{(b)}) = f_{\psi}(x, \sigma, y_\text{robot})$
            \State Decode risk-biased predictions $\hat{y}_k = g_\theta(x, \hat{z}_k^{(b)})$
        \EndFor
        \State \smash{Compute expected cost $\hat{r} = \tfrac{1}{K_2}\sum_{k=1}^{K_2} J^{y_{\text{robot}}}(\hat{y}_k)$}
        \State Compute risk loss $L_\text{risk} = \rho(\hat{r} - r)$ and prior loss $L_\text{prior} = \text{KL}\left(\mathcal{N}(\mu^{(b)}, \Sigma^{(b)}) || \mathcal{N}(\mu|_x, \Sigma|_x) \right) $
        \OUTPUT Loss value $\alpha L_\text{risk} + \beta L_\text{prior}$ to train $\psi$ ($\theta$ and $\phi_1$ are fixed)
    \end{algorithmic}
    \label{algo: risk-bias loss}
\end{algorithm}
}

Algorithm \ref{algo: risk-bias loss} lays out the procedure for training our proposed risk-aware prediction. It relies on a fully trained CVAE with the encoder $f_{\phi_1}: x \to (\mu|_x, \Sigma|_x)$ and decoder $g_\theta: x, z \to y$ that fits the distribution of $Y|_x$ from a dataset. We train a new latent-biasing encoder $f_{\psi}: x, \sigma, y_\text{robot} \to (\mu^{(b)}, \Sigma^{(b)})$ to bias the latent distribution while keeping the rest of the CVAE fixed. The risk-level $\sigma$ is randomly sampled on $[0, 1]$ during training and chosen by the user at test time.

\section{Experiments}
\label{sec: experiments}

\subsection{Biasing forecasts in a didactic scenario}
\label{sec:didactic_biasing}
\begin{figure}[h]
    \vspace{-12pt}
    \centering
    \adjustbox{width=\columnwidth, trim=1.4cm 0cm 1cm 0cm, clip}{%
        \includegraphics{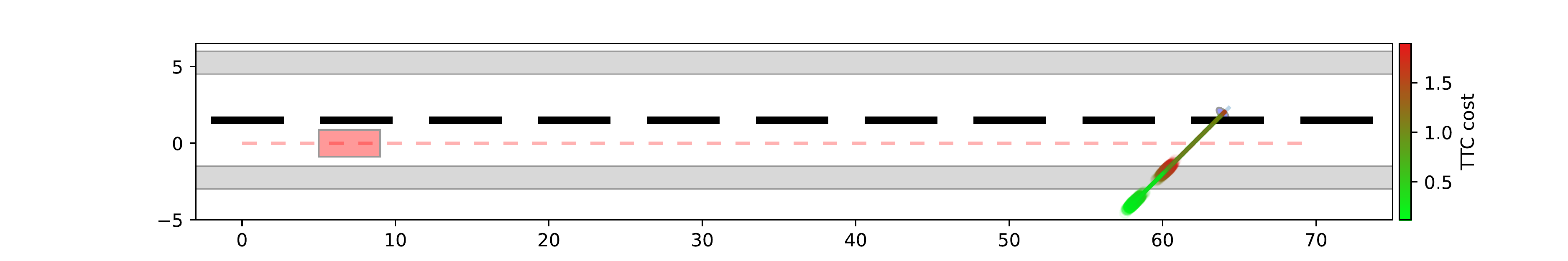}
    }
    \caption{Top-down view of a simulated scene. The robot in red moves left to right down the road as a pedestrian in blue is crossing. The color of the depicted pedestrian trajectory samples indicates their corresponding Time-To-Collision (TTC) cost for the robot. The slow mode in red is more costly than the fast mode in green.}
    \label{fig: road_scene}
    \vspace{6pt}
\end{figure}

\begin{figure}[h]
    \vspace{-12pt}
    \centering
    \begin{subfigure}[b]{0.3\columnwidth}
        \centering
        \adjustbox{width=0.9\columnwidth, trim=0.18cm 0.2cm 2.5cm 0.1cm, clip}{%
            \includegraphics{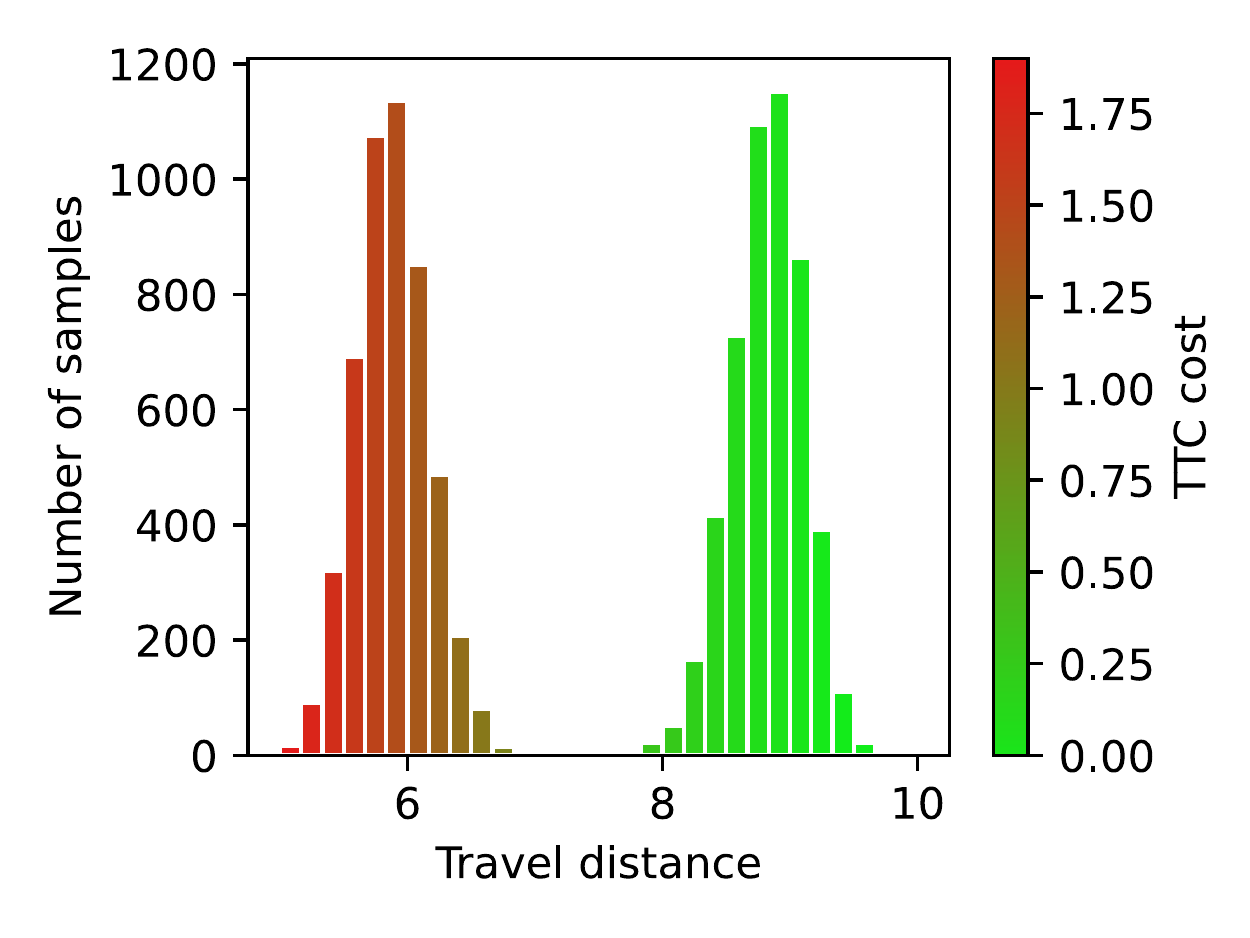}
        }
        \caption{Ground truth data distribution.}
        \label{subfig: data distrib}
    \end{subfigure}
    \hfill
    \begin{subfigure}[b]{0.3\columnwidth}
        \centering
        \adjustbox{width=0.9\columnwidth, trim=0.18cm 0.2cm 2.5cm 0.1cm, clip}{%
            \includegraphics{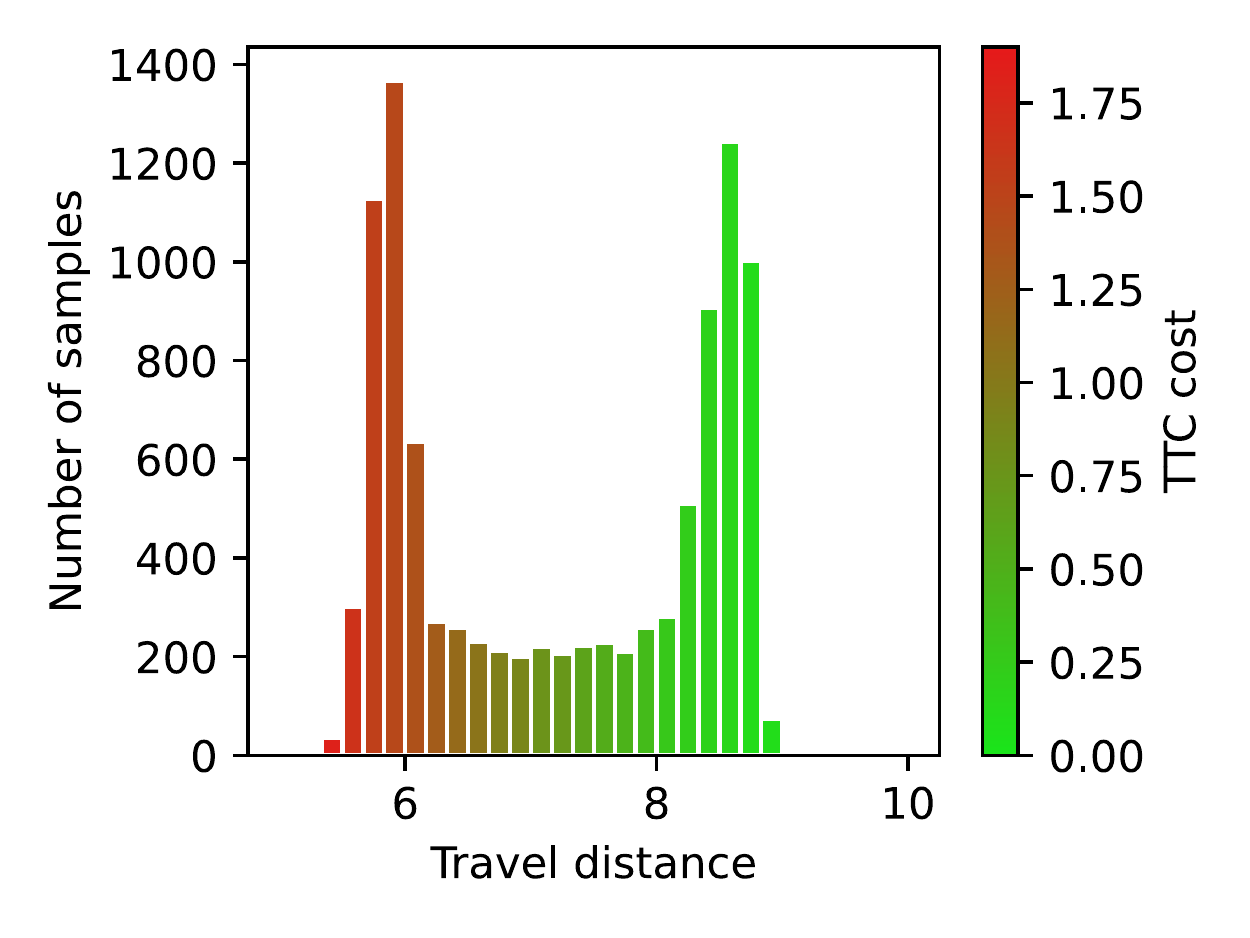}
        }
        \caption{Risk-neutral prediction.}
        \label{subfig: predicted distrib}
    \end{subfigure}
    \hfill
    \begin{subfigure}[b]{0.36\columnwidth}
        \centering
        \adjustbox{width=\columnwidth, trim=0.18cm 0.2cm 0.0cm 0.1cm, clip}{%
            \includegraphics{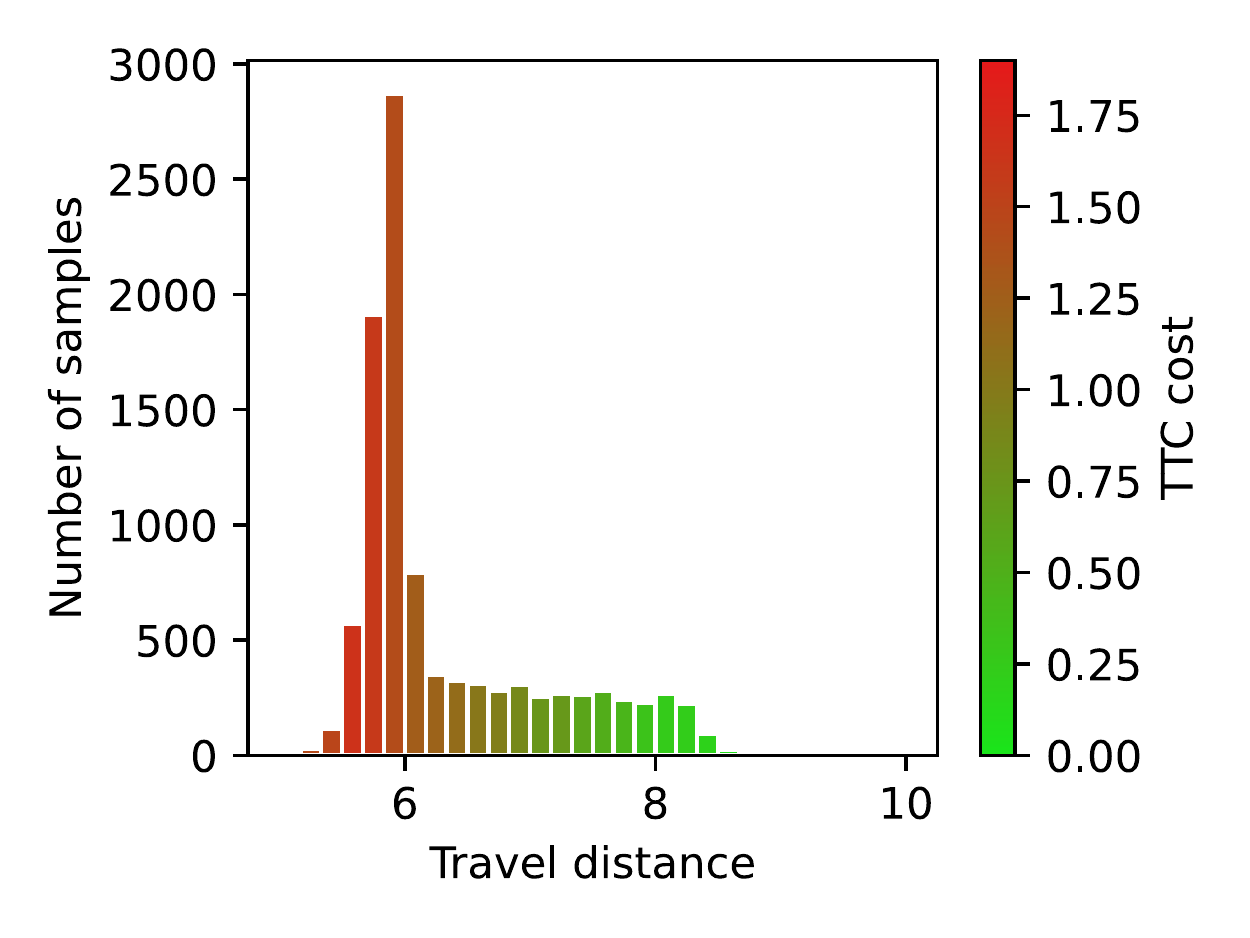}
        }
        \caption{Risk-biased prediction $\sigma=0.95$.}
        \label{subfig: safer slow distrib}
    \end{subfigure}
    \hfill
    \vspace{10pt}
    \caption{Histograms of the pedestrian travel distances at the end of the 5-second episode in the defined scene. Each bar is colored with the average Time-To-Collision (TTC) cost of the bin.}
    \label{fig:hist_travel_distance}
\end{figure}

We created the didactic simulation environment in \cref{fig: road_scene} where a red robot drives at constant speed along a straight road with a stochastic pedestrian.
The pedestrian either walks slowly or quickly, yielding a bimodal distribution over their travel distance.
We collected a dataset in this environment where the initial position and orientation of the pedestrian are set at random. We used it to train a risk-biased CVAE model according to the method presented in sections \ref{sec: problem_formulation} and \ref{sec: implementation}.
\cref{subfig: predicted distrib} shows the risk-neutral prediction ($\sigma = 0$) of the pedestrian's travel distance in a specific scene.
As can be seen, the model captures both of the equally-likely modes.
In contrast, in the risk-biased case ($\sigma = 0.95$), the model predicts the slower mode with much greater frequency because, in this scene, if the pedestrian walks slowly it will collide with the robot.
If, alternatively, the pedestrian walks quickly, the vehicle will pass behind it safely without collision.
In other words, the risk-biased model pessimistically predicts collisions with a greater probability than does the risk-neutral model. With $\sigma = 0.95$, pessimism is so high that the safer mode falls in the tail of the distribution. In supplemental \ref{sec:latent_explore} we explore the latent representation of this model.


\subsection{Planning with a biased prediction}
\label{sec: didactic_biasing_with_mpc}
\begin{figure}[t]
    \vspace{-12pt}
    \centering
    \adjustbox{width=1.0\columnwidth, trim=2cm 0cm 2cm 0cm, clip}{%
    \includegraphics{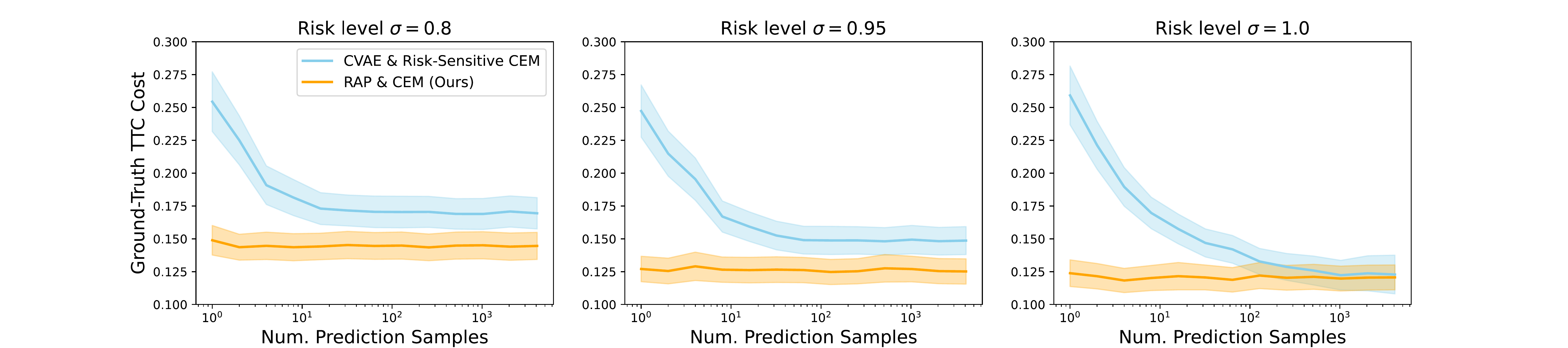}
    }
    \caption{Ground-truth TTC cost of the optimized $y_{\text{robot}}^*$, averaged over 500 episodes (lower the better). Ribbons show 95\% confidence intervals of the mean. Our risk-biased predictor (RAP) coupled with CEM consistently achieves low cost regardless of the number of prediction samples that CEM draws online from the predictor.}
    \label{fig: stats_mpc}
\end{figure}
The previous experiment demonstrates the ability of our approach to bias predictions towards dangerous outcomes.
The experiment in this section evaluates whether this ability can benefit online planning using a model-based trajectory optimization algorithm.
In this setting, predictions are used to evaluate the risk of various candidate robot trajectories, $y_{\text{robot}}$, in order to select the best one.

We generated a new dataset in which the robot's initial speed and per-timestep accelerations are sampled randomly, as opposed to the constant velocity model used previously.
This variation ensures that the robot trajectories generated by the planner are within the training distribution.
We modified the biasing encoder to account for the changing robot trajectories by adding $y_{\text{robot}}$ to its inputs. This allows our new model to achieve pessimistic forecasting with respect to a particular $y_{\text{robot}}$.

The online planner controls the longitudinal acceleration of the robot, which is modeled as a double integrator system.
We employed the cross entropy method (CEM)~\cite{chua2018deep, nagabandi2020deep} as the underlying optimization algorithm.
CEM is a stochastic optimization method that locally optimizes $y_{\text{robot}}$ based on a given initial $y_{\text{robot}}^{\text{init}}$.
In each episode, CEM first draws $n_{\text{samples}}$ risk-biased prediction samples given $y_{\text{robot}}^{\text{init}}$ and observed pedestrian motion $x$, and uses the samples to produce a locally optimal $y_{\text{robot}}^*$.
This process involves a linear complexity in $n_{\text{samples}}$.
We incorporated a quadratic trajectory tracking cost in the planner's objective so that the robot is encouraged to continue moving at a constant speed when the risk of collision is deemed negligible.
See supplementary material \ref{sec:experimental_details_mpc} for details.

We evaluated the performance of the combined risk-biased predictor and CEM planner across 500 episodes.
\cref{fig: stats_mpc} shows that CEM using the risk-biased predictor consistently produces $y_{\text{robot}}^*$ with low Time-To-Collision (TTC) cost values, even when sampling few trajectories.
We compared with a baseline approach in which a risk-sensitive version of CEM performs planning using an unbiased CVAE predictor.
We obtained this risk-sensitive planner by replacing the Monte Carlo expectation with the CVaR estimator~\cite{hong2014monte}.
This baseline is an instance of the conventional risk-sensitive planning with data-driven human motion forecasting~\cite{nishimura2020risk, novin2021risk}, which evaluates the risk within the planner rather than in the predictor.
\cref{fig: stats_mpc} shows significantly higher TTC cost of $y_{\text{robot}}^*$ for the baseline when CEM uses fewer than 16 samples. This is because the collision risk is underestimated with few samples, and thus the planner over-optimistically optimizes trajectory tracking to the detriment of safety.
\subsection{Robustness to out-of-distribution pedestrian behavior}
\label{sec:distributional_robustness_test}
For this last didactic experiment, we evaluated the distributional robustness of the overall prediction-planning pipeline to a test-time change of the pedestrian stochastic behavior model.
Specifically, we used the same dataset and planner as in Section \ref{sec: didactic_biasing_with_mpc}, but we reduced the overall average speed of the pedestrian by 25\% \textit{only at test time, after training}.
Other factors such as bi-modality were kept the same as at training time.
From the robot's perspective, reducing the average speed of the pedestrian in the test scenario (as exemplified in Fig. \ref{fig: road_scene}) results in an adversarial distribution shift.
We studied how robust the risk-aware robot is under this out-of-distribution pedestrian behavior that the predictor did not witness during training.

\begin{table}[t]
\centering 
\caption{
Ground-truth TTC cost of the optimized $y^*_{\text{robot}}$ under different test-time pedestrian behaviors, averaged over 500 episodes (lower the better).
}
\label{tab:distributional_robustness}
\resizebox{\textwidth}{!}{
\begin{tabular}{lcllcc}
\hline
Predictive Model & \# Prediction Samples & Planner & $\sigma$ & Pedestrian Behavior & TTC Cost\\
\hline
Unbiased CVAE & 64 & Risk-Neutral CEM & \scriptsize{NA} & same as training & $0.23$ \scriptsize{$\pm 0.01$} \\
\hline
Unbiased CVAE & 64 & Risk-Neutral CEM & \scriptsize{NA} & 25\% reduced speed & $0.44$ \scriptsize{$\pm 0.02$} \\
\hline
Unbiased CVAE & 64 & Risk-Sensitive CEM & 0.95 & 25\% reduced speed & $0.37$ \scriptsize{$\pm 0.02$} \\
Biased CVAE (\modelacronym) & 64 & Risk-Neutral CEM  & 0.95 & 25\% reduced speed & $0.34$ \scriptsize{$\pm 0.02$} \\
\hline
Unbiased CVAE & 1 & Risk-Sensitive CEM & 0.95 & 25\% reduced speed & $0.46$ \scriptsize{$\pm 0.02$} \\
Biased CVAE (\modelacronym) & 1 & Risk-Neutral CEM & 0.95 & 25\% reduced speed & $0.34$ \scriptsize{$\pm 0.02$} \\
\hline
\end{tabular}
}
\vspace{-12pt}
\end{table}

Table \ref{tab:distributional_robustness} summarizes the results of this experiment.
The first row shows the nominal in-distribution risk-neutral case.
The second row shows that the risk-neutral robot is not robust at all to the distribution shift, resulting in the doubling of the nominal TTC cost.
This is expected from an autonomy stack without risk-awareness.
Rows 3 and 4 suggest that risk-awareness improves robustness given sufficient samples from the predictor.
Finally, the last two rows show that our proposed framework remains robust to the distribution shift \textit{even with a single prediction sample}, whereas the conventional risk-sensitive prediction-planning approach (row 5) does not show robustness any more.
This demonstrates that, when the computation budget for online planning is limited, distributional robustness is better achieved within the predictor rather than in the planner.

\subsection{Application to real-world data}
\label{sec: real_world_biasing}
\begin{figure}[t]
    \centering
    \begin{subfigure}[b]{0.24\linewidth}
        \centering
        \includegraphics[width=\linewidth, trim={5cm 8.5cm 6cm 7.5cm},clip]{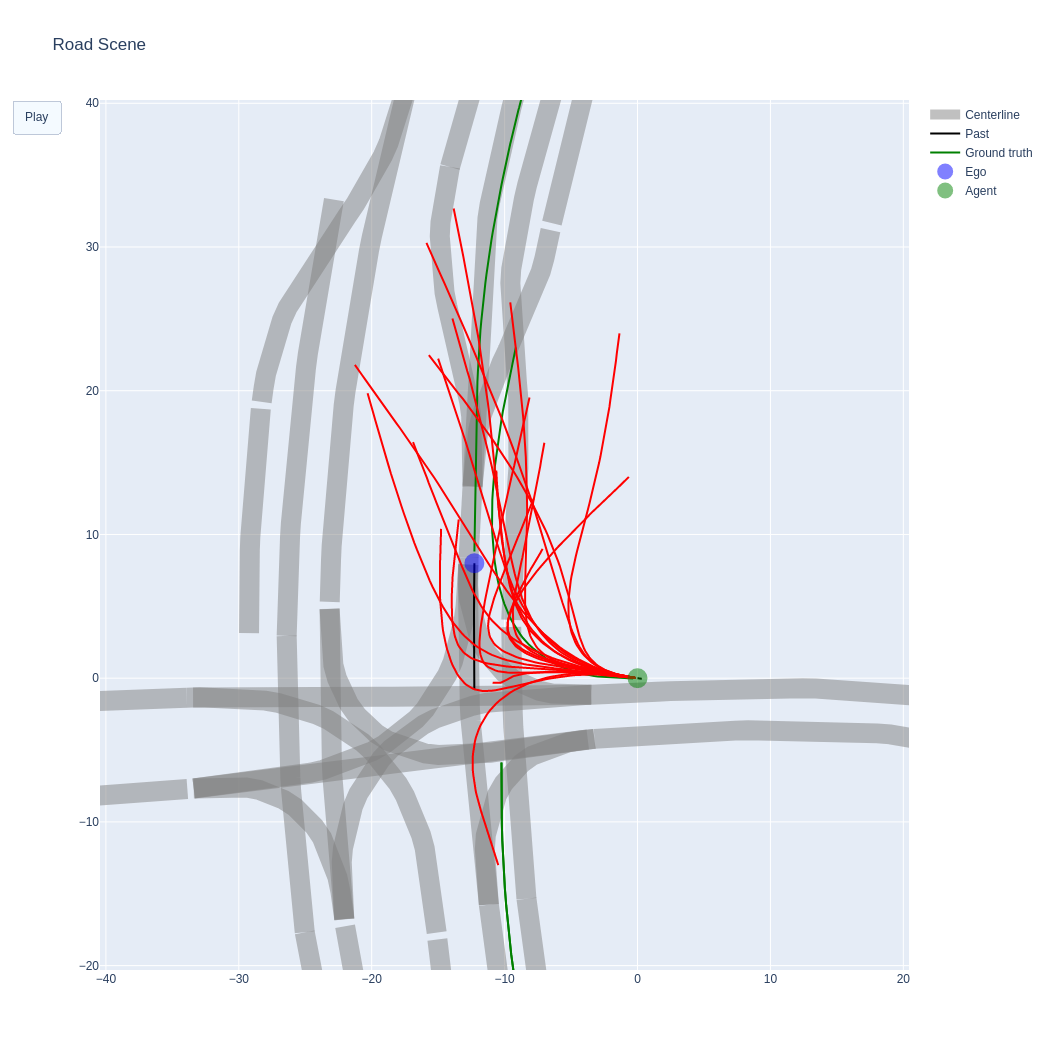}
        \caption{Risk-neutral}
    \end{subfigure}
    \hfill
    \begin{subfigure}[b]{0.24\linewidth}
        \centering
        \includegraphics[width=\linewidth, trim={5.5cm 8cm 5.5cm 8cm},clip]{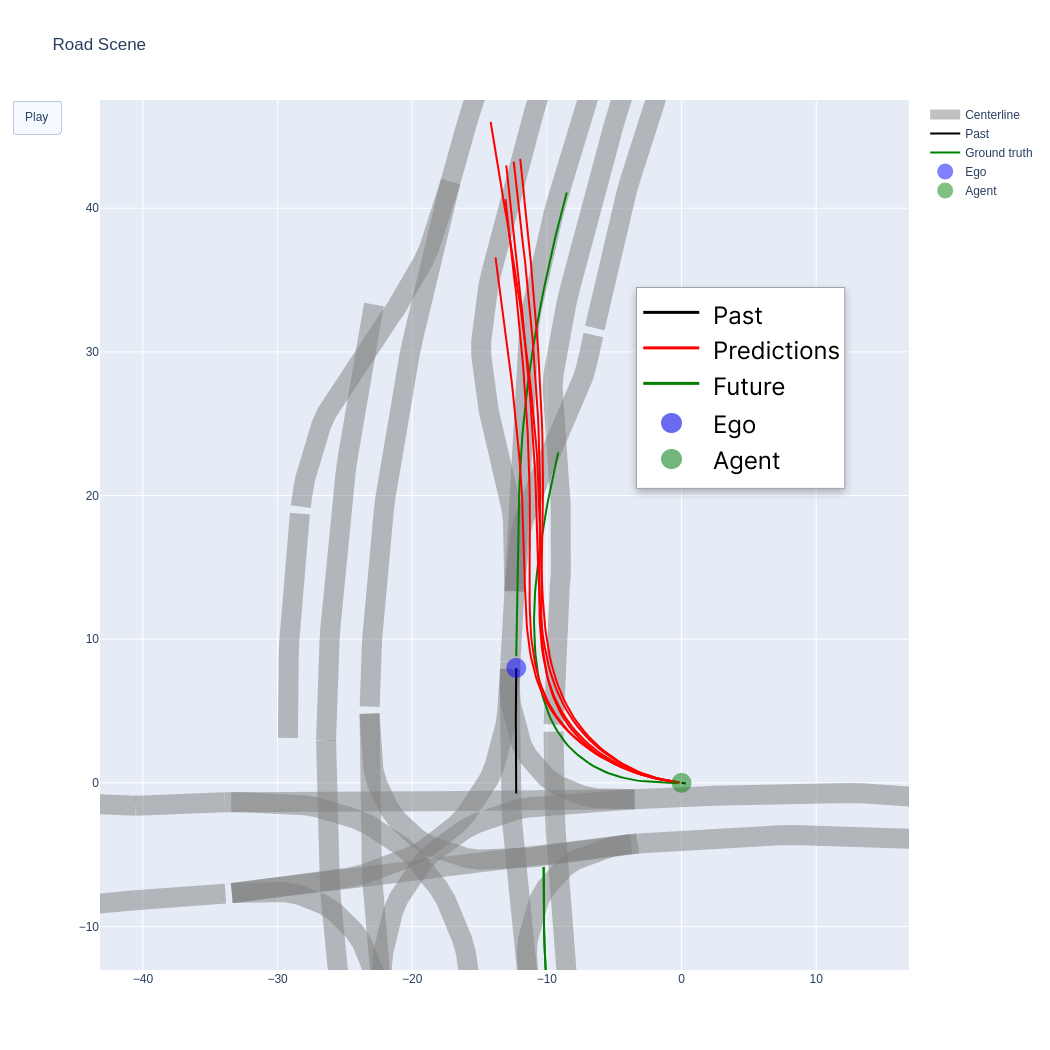}
        \caption{\textbf{Risk-aware}, $\sigma=0.95$}
    \end{subfigure}
     \hfill
    \begin{subfigure}[b]{0.24\linewidth}
        \centering
        \includegraphics[width=\linewidth, trim={2cm 10cm 6cm 4cm},clip]{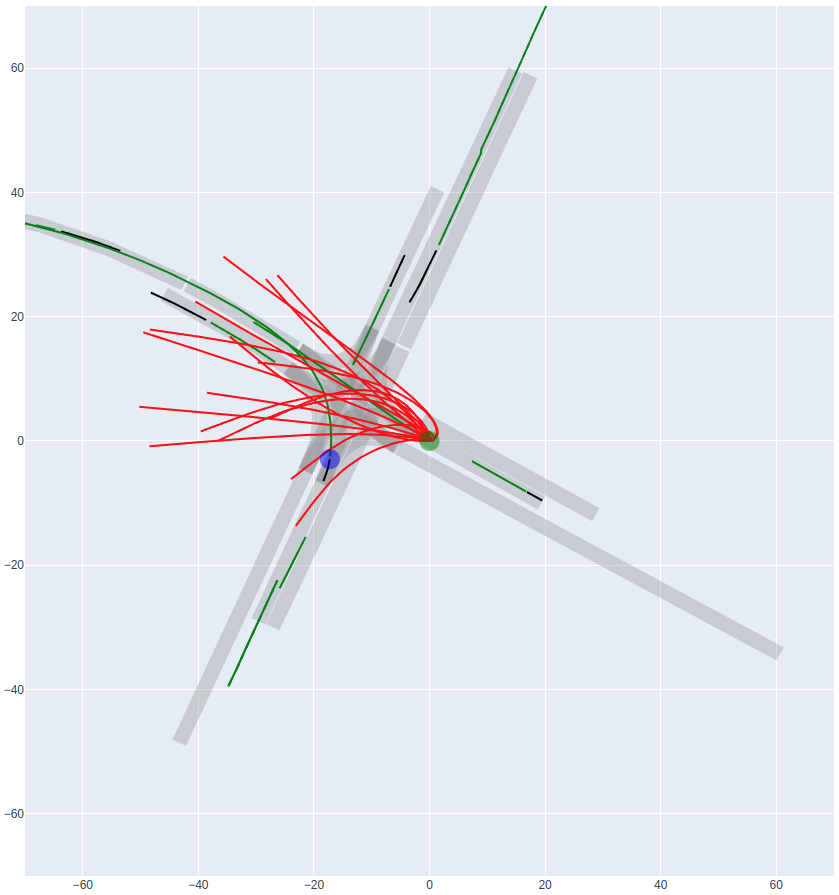}
        \caption{Risk-neutral}
    \end{subfigure}
    \hfill
    \begin{subfigure}[b]{0.24\linewidth}
        \centering
        \includegraphics[width=\linewidth, trim={2cm 10cm 6cm 4cm},clip]{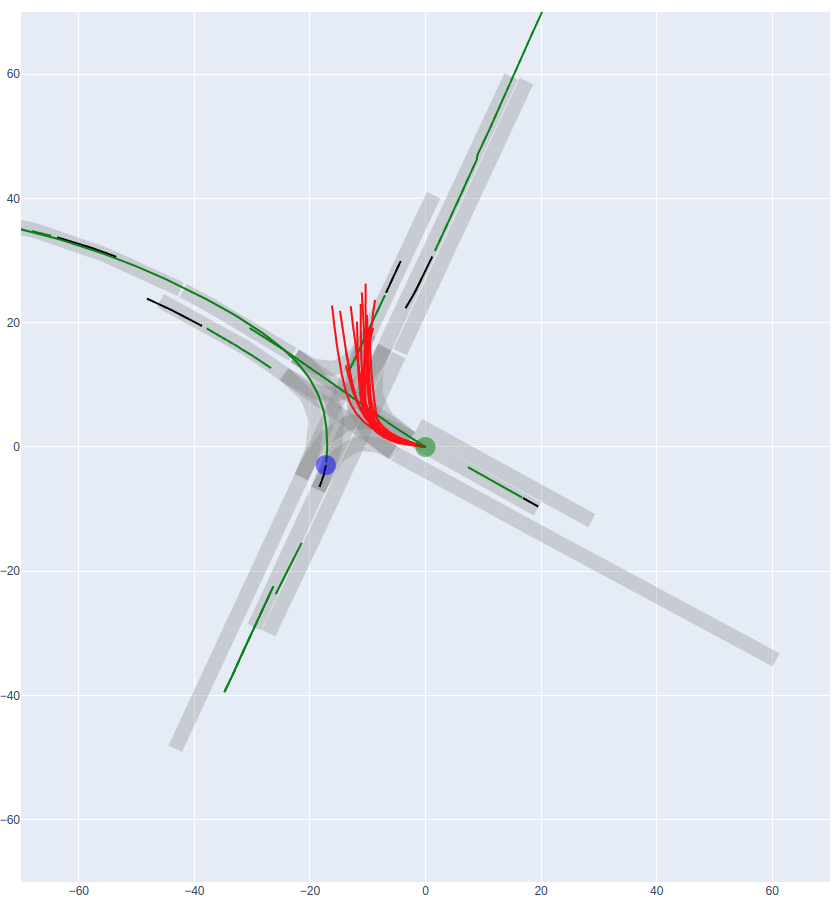}
        \caption{\textbf{Risk-aware}, $\sigma=0.95$}
        \label{fig:qualitative_waymo_failure}
    \end{subfigure}\\
    \vspace{4mm}
    \caption{Visualization of WOMD scenes and forecasts. The lane centerlines are represented in gray, the past observations in black, the ground truth futures in green, and the 16 forecast samples in red. The ego is in blue and the agent to predict in green.
    In figure \textbf{(b)} the risk-aware forecasts turns towards the ego. In figure \textbf{(d)} the risk-aware forecasts would be costly if the ego went straight ahead. This is a failure case where the forecasts are biased towards an expected ego trajectory that did not occur. 
    }
    \label{fig:qualitative_waymo}
\end{figure}

We applied our risk-biased forecasting framework to real-world data from the Waymo Open Motion Dataset (WOMD) \cite{ettinger2021large}.
Following a similar approach to~\cite{huang2021tip}, we selected the annotated scenarios that cover interesting interactions between two agents. We randomly selected one of the two interacting agent as the ego and the other as the agent to predict. We input other agent tracks and the map as additional conditioning information to account for the interaction with the environment and the other agents. Then, we trained a biased CVAE model as described in Section \ref{sec: implementation}.
In this experiment we only conditioned the biased-encoder on the ego past trajectory, not its future trajectory, in order to avoid ground truth information leakage.
This means that the biased-encoder is making an implicit forecast of the ego future, which leads to the failure mode presented in \cref{fig:qualitative_waymo_failure} wherein the wrong implicit forecast leads to an incorrectly-biased distribution.

\begin{table}[t]
\centering 
\caption{Motion forecasting error and risk estimation error on the WOMD validation set. 
\textbf{minFDE (K)}: minimum final displacement error over K samples,
\textbf{risk error (K)}: mean value of the signed difference between the average cost of the biased forecasts over K samples and the risk estimation using the unbiased forecasts,
\textbf{risk $|$error$|$ (K)}: mean value of the absolute values of the risk estimation error.}  
\label{tab:waymo-results}
\resizebox{0.9\textwidth}{!}{
\begin{tabular}{lccccc}
\hline
Predictive Model & $\sigma$ & minFDE(16) & FDE (1) & Risk error (4) & Risk $|$error$|$ (4) \\
\hline
Unbiased CVAE & \scriptsize{NA} &$3.82$ \scriptsize{$\pm 0.04$}& $13.06$ \scriptsize{$\pm 0.01$} & \scriptsize{NA} & \scriptsize{NA} \\
\hline
Biased CVAE (\modelacronym) & 0 & $3.81$ \scriptsize{$\pm 0.04$}& $13.07$ \scriptsize{$\pm 0.02$}&  $0.00$ \scriptsize{$\pm 0.00$}& $0.12$ \scriptsize{$\pm 0.00$}\\
Biased CVAE (\modelacronym) & 0.3 & $4.32$ \scriptsize{$\pm 0.05$}& $11.89$ \scriptsize{$\pm 0.02$}& $0.02$ \scriptsize{$\pm 0.00$}& $0.13$ \scriptsize{$\pm 0.00$}\\
Biased CVAE (\modelacronym) & 0.5 & $5.32$ \scriptsize{$\pm 0.06$}& $12.05$ \scriptsize{$\pm 0.02$}&  $0.02$ \scriptsize{$\pm 0.00$}& $0.16$ \scriptsize{$\pm 0.00$}\\
Biased CVAE (\modelacronym) & 0.8 & $7.78$ \scriptsize{$\pm 0.07$}& $13.53$ \scriptsize{$\pm 0.02$}&  $0.01$ \scriptsize{$\pm 0.00$}& $0.26$ \scriptsize{$\pm 0.00$}\\
Biased CVAE (\modelacronym) & 0.95 & $10.13$ \scriptsize{$\pm 0.09$}& $15.33$ \scriptsize{$\pm 0.02$}&  $0.03$ \scriptsize{$\pm 0.01$}& $0.43$ \scriptsize{$\pm 0.00$}\\
Biased CVAE (\modelacronym) & 1 & $11.58$ \scriptsize{$\pm 0.09$}& $16.29$ \scriptsize{$\pm 0.02$}& $-0.22$ \scriptsize{$\pm 0.01$}& $0.60$ \scriptsize{$\pm 0.01$}\\
\hline
\end{tabular}
}
\vspace{-12pt}
\end{table}

Table \ref{tab:waymo-results} shows the results of this experiment. First note the large difference between the minFDE and FDE values of the unbiased model, which illustrates that the predictions are diverse. This is qualitatively supported by \cref{fig:qualitative_waymo}, which shows a wide diversity of predicted trajectories. 
As expected, our biased CVAE model (\modelacronym) with a risk level $\sigma = 0$ shows results that match the ones from the unbiased model.
As the risk-level increases, the predictions increasingly differ from those of the unbiased distribution as well as the ground truth trajectory. This is reflected by the larger minFDE and FDE values. At $\sigma=1$, the biased prediction distribution collapses to the mode that the model estimates to be the most costly, yielding minFDE close to FDE.
The risk estimation error is the average difference of the mean \emph{cost} under the biased prediction and the \emph{risk} estimated using a large number of samples from the unbiased prediction. Its mean value shows the risk estimation bias of the proposed approach while its mean \textit{absolute} value shown in the next column shows the average error that is made in either direction. Up to $\sigma=0.95$, the risk estimation is nearly unbiased. At $\sigma=1$, the risk estimation is slightly under-estimated. 
Usage of real data in this section limits our ability to provide accurate safety statistics for our approach. Instead, we provide extensive qualitative results.
Section \ref{sec: experiment_details} of the supplement gives additional experimental details and results. We also provide extra experiments, figures and animations on our project website\footnote{\url{https://sites.google.com/view/corl-risk/home}}. Finally, the model can be tested directly on several hundred samples, with any risk-level, on our online demo\footnote{\url{https://huggingface.co/spaces/TRI-ML/risk_biased_prediction}}.


\section{Limitations}
\label{sec: limitations}

A first limitation of our approach is that the constrained optimization problem~\eqref{eq:optimization_problem} is difficult to solve in practice due to challenges presented in Section 4.
The constraint relaxation and the neural network optimization method yield a sub-optimal risk-aware predictor that may still underestimate risk when the risk-sensitivity level is close to 1. %
Therefore, our method might be inadequate when an extremely conservative behavior is desired, or in the case of extremely low probability but catastrophic events.
Second, \emph{in the real data application}, our risk-aware prediction is not conditioned on a specific robot plan which may lead the forecast to collapse onto a mode that is not the most critical.
Finally, we only forecast marginal agent behavior instead of jointly predicting the behaviors of several agents in the scene. This neglects potential risk-avoiding interactions in the future and leads to overly pessimistic biased forecasts.

\section{Conclusion}
\label{sec: conclusion}
This paper proposes a risk-aware trajectory forecasting method for robust planning in human-robot interaction problems.
We present a novel framework to learn a pessimistic distribution offline that simplifies online risk evaluation to expected cost estimation.
Our experimental results show that this method leads to safe robot plans with reduced sample complexity.
We additionally demonstrate the effectiveness of our approach in real-world scenarios with low risk-estimation error and strong qualitative results.
In future work, we intend to evaluate our approach in a realistic simulator, and also improve the accuracy of risk estimation on real-world data by conditioning biased prediction on potential robot plans.


\clearpage


\bibliography{references.bib}

\clearpage
\appendix

\section{Analysis of the Risk-Constrained Minimization of the KL-Divergence}
\label{sec: proofs}

\begin{prop}
   Let $\mathcal{Y}$ be a connected set (in practice $\mathbb{R}^n$), $J^\pi:\ \mathcal{Y} \rightarrow \mathbb{R}^+$ a continuous cost function, $p$ the density function of the random vector $Y$ on $\mathcal{Y}$, $\sigma \in \mathbb{R}$ a risk-level, and $\mathcal{R}_p(J^\pi(Y),\sigma) \in \mathbb{R}$ a risk measure such that  $\inf(J^\pi)<\mathcal{R}_p(J^\pi(Y),\sigma)<\sup(J^\pi)$.
   Then, there exists at least one density function $q_\psi$ of a random vector $Y|_{\sigma}$ on $\mathcal{Y}$ such that $\mathbb{E}_{q_{\psi}}[J^\pi(Y|_{\sigma})] = \mathcal{R}_p(J^\pi(Y),\sigma)$.
\label{prop: existence}
\end{prop}
\begin{proof}
    For any value $j^* \in ]\inf(J^\pi), \sup(J^\pi)[$, the intermediate value theorem states that there exist $y^* \in \mathcal{Y}$ such that $J^\pi(y^*)=j^*$. In particular, for any value of $\sigma$, we can choose $j^*_{\sigma} = \mathcal{R}_p(J^\pi(Y),\sigma)$.
    Let us now define the density function $q^*_{\sigma}: y \to \delta(y-y^*)$, with $\delta$ the Dirac delta density function.
    We obtain the equality $\mathbb{E}_{q^*_{\sigma}}[J(Y|_{\sigma})] = j^* = \mathcal{R}_p(J(Y),\sigma)$, which proves that there is always at least one solution to equation \eqref{eq:risk_constraint}.
\end{proof}

\begin{prop}
    Let $\Delta_{\sigma} = \left\{ q \text{ s.t. } \mathbb{E}_{q}[J^\pi(Y)] = \mathcal{R}_p(J,\sigma)\right\}$. Let $q_\psi$ be the density function on $\mathcal{Y}$ that is parameterized by $\psi$ and $p$ the density function on $\mathcal{Y}$ from which the dataset is sampled.
    
    Then, there exists a unique density function $ q_{\psi^*} \in \Delta_{\sigma}$ that minimizes $\text{KL}(q_\psi||p)$. 
\label{prop: unicity}
\end{prop}
\begin{proof}
    Let $q_1, q_2 \in \Delta_{\sigma}$ that both minimize $\text{KL}(q_{\psi}||p)$ for a given $p$:
    \[
        \text{KL}(q_{1}||p) = \text{KL}(q_{2}||p) = \min_{q_\psi \in \mathcal{R}_{\sigma}}\big(\text{KL}(q_\psi || p)\big)
    \]
    
    As a first step for this proof, we show that for any $\alpha \in [0, 1]$, we have
    $\alpha q_{1} + (1-\alpha)q_{2} \in \Delta_{\sigma}$.
    Then, given that the function $q \rightarrow \text{KL}(q||p)$ is strictly convex, we use the equality case of Jensen's inequality to show that $q_1 = q_2$ almost everywhere, which proves the uniqueness. 
    
    \begin{align*}
        \mathbb{E}_{\alpha q_{1} + (1-\alpha)q_{2}}[J] &= \int J(y)(\alpha q_{1}(y) + (1-\alpha)q_{2}(y))dy\\
        &= \alpha \int J(y)\alpha q_{1}(y)dx + (1-\alpha)\int J(y)q_{2}(y)dy\\
        &= \alpha\mathbb{E}_{q_{1}}[J] + (1-\alpha)\mathbb{E}_{q_{2}}[J]\\
        &= \alpha R_p(J,\sigma) + (1-\alpha)R_p(J,\sigma)\\
        &= R_p(J,\sigma)
    \end{align*}
    Therefore, for any $\alpha \in [0, 1]$, $\alpha q_{1} + (1-\alpha)q_{2} \in \Delta_{\sigma}$.

    The Jensen's inequality with $\text{KL}(\cdot||p)$ gives us:
    \[
        \text{KL}(\alpha q_{1} + (1-\alpha)q_{2}||p) \leq \alpha\text{KL}(q_{1}||p) + (1-\alpha)\text{KL}(q_{2}||p)
    \]
    From our definition of $q_1$ and $q_2$:
     \begin{align*}
         \text{KL}(q_{1}||p) = \text{KL}(q_{2}||p) = \min_{q \in \Delta_{\sigma}}\big(\text{KL}(q||p)\big)\\ = \alpha\text{KL}(q_{1}||p) + (1-\alpha)\text{KL}(q_{2}||p)
     \end{align*}
     
    Since $\alpha q_{1} + (1-\alpha)q_{2} \in \Delta_{\sigma}$ and is lower or equal to $\min_{q \in \Delta_{\sigma}}\big(\text{KL}(q||p)\big)$, it is equal.
    Therefore, there is equality in the Jensen's inequality with a strictly convex function. This means that $q_{1} = q_{2}$ almost everywhere and concludes our proof.
    
\end{proof}

 In this proof, the uniqueness is established for $q$ minimizing $\text{KL}(q||p)$ in the data (i.e., trajectory) space, which is not what we do in practice. In \eqref{eq:optimization_problem} we minimize the KL-divergence from the prior in the latent space.
    
    One might be tempted to define $l_1$ and $l_2$ as two biased density functions in the latent space such that $l_1 = q_{1} \circ g_{\theta}$ and $l_2 = q_{2} \circ g_{\theta}$.
    However, such $l_1$ and $l_2$ might not be density functions at all because they would not always integrate to one.
    We need to assume that $g_{\theta}$ is volume preserving for $l_1$ and $l_2$ to be well defined.
    
\begin{definition}
        A differentiable function $g: \mathbb{R}^n \to \mathbb{R}^n$ is \textbf{volume-preserving} if $|\det(J_{g}(z))|=1$ for all $z \in \mathbb{R}^n$, where $J_{g}$ is the Jacobian of $g$.
\label{def:volume_preserving}
\end{definition}
    Let us make two remarks about volume preservation in our application:
    First, it requires that $z$ and $y$ share the same dimension ($z$ and $y$ $\in \mathbb{R}^n$) for the determinant of the Jacobian to be defined. This means that the information bottleneck of the CVAE is not straightforward to enforce.
    Second, with the CVAE assumptions, the elements of the latent variable should be independent, thus $J_{g_{\theta}}$ should be diagonal and $|\det(J_{g_{\theta}})|= |\nabla \cdot g_{\theta}|$.

\begin{prop}
    Let $q_\psi$ be a latent density function, $\rho$ a prior latent density function and $g_{\theta}$ (the decoder) be a volume-preserving function, all defined on $\mathbb{R}^n$. 
    We suppose that $g_\theta$ fits the dataset such that $ \mathcal{R}_p(J ,\sigma) = \mathcal{R}_\rho(J \circ g_\theta ,\sigma)$, with $p$ the data distribution on $\mathcal{Y} = \mathbb{R}^n$.
    Let us define the density function $l_\psi = q_\psi \circ g_\theta$ on $\mathbb{R}^n$.
    Finally, we define the constraint domain $\Delta_{\sigma} = \left\{ l \text{ s.t. } \mathbb{E}_{l}[J \circ g_\theta] = \mathcal{R}_p(J, \sigma) = \mathcal{R}_\rho(J \circ g_\theta, \sigma) \right\}$. 

    Then, there exists a unique density function $ l_{\psi^*} \in \Delta_{\sigma}$ that minimizes $\text{KL}(l_\psi||\rho)$. 
\label{prop: unicity_vp}
\end{prop}
    
\begin{proof}
    Let $l_1, l_2  \in \Delta_{\sigma}$ be two latent density functions that minimize the KL-divergence with the prior $\rho$:
    \[
        \text{KL}(l_{1}||\rho) = \text{KL}(l_{2}||\rho) = \min_{l \in \Delta_{\sigma}}\big(\text{KL}(l||\rho)\big)
    \]
    
    Let us show that $\alpha l_{1} + (1-\alpha)l_{2} \in \Delta_{\sigma}$:
    \begin{align*}
         R_p(J,\sigma) &= \mathbb{E}_{\alpha q_{1} + (1-\alpha)q_{2}}[J] \\
        &= \int J(y)(\alpha q_{1}(y) + (1-\alpha)q_{2}(y))dy\\
        &=  \int J(g_{\theta}(z))(\alpha q_{1}(g_{\theta}(z)) + (1-\alpha)q_{2}(g_{\theta}(z)))|\det(J_{g_{\theta}}(z))| dz\\
        &=  \int J(g_{\theta}(z))(\alpha l_{1}(z) + (1-\alpha)l_{2}(z))dz\\
        &=  \mathbb{E}_{\alpha l_{1} + (1-\alpha)l_{2}}[J\circ g_{\theta}]
    \end{align*}
    And thus $\alpha l_{1} + (1-\alpha)l_{2} \in \Delta_{\sigma}$. The rest of the proof of \cref{prop: unicity} holds.

\end{proof}

\section{Details about the neural network architectures}
\label{sec: architecture}

\subsection{Learning to forecast}

In section \ref{sec: experiments}, we perform experiments that rely on specific implementations of trajectory forecasting models. We use two different architectures: the first is a small MLP-based model used in our proof-of-concept experiments (Sections \ref{sec:didactic_biasing}, \ref{sec: didactic_biasing_with_mpc}, and \ref{sec:distributional_robustness_test}), and the second resembles state-of-the-art forecasting architectures used in challenging, real-world scenarios (Section \ref{sec: real_world_biasing}).
As described in Section \ref{sec: problem_formulation}, our method relies on a latent space to bias the forecasts.
We use CVAEs in this work, however, our method can be applied more broadly to any forecasting model that conditions on a latent sample.

The small model is composed of two multi-layer perceptron (MLP) encoders and one MLP decoder.
The inference encoder takes the past trajectory $x$ and outputs the parameters of a Normal distribution $\mu|_x$ and $\log(\operatorname{diag}(\Sigma|_x))$.
The posterior encoder takes the whole past and future trajectory $x, y$ and outputs the parameters of a Normal distribution $\mu|_{x,y}$ and $\log(\operatorname{diag}(\Sigma|_{x,y}))$.
Finally, the decoder takes the past trajectory $x$ and a latent sample $z$ and outputs a prediction $y$.
The second architecture is designed to model the interactions with both the other agents and map elements. 
It takes a similar form as the small model, but with additional context inputs and larger hidden dimensions.
The social and map interactions are accounted for with a modified multi-context gating block \cite{varadarajan2021multipath++}.

\subsection{Model architecture}

The first architecture is a simple model used in our proof-of-concept experiments. It is composed of three multi-layer perceptron (MLP) encoders and one MLP decoder:
\begin{itemize}
    \item The inference encoder takes the past trajectory $x$ and outputs the parameters of a Normal distribution $\mu|_x$ and $\log(\operatorname{diag}(\Sigma|_x))$.
    \item The posterior encoder  takes the whole past and future trajectory $x, y$ and outputs the parameters of a Normal distribution $\mu|_{x,y}$ and $\log(\operatorname{diag}(\Sigma|_{x,y}))$.
    \item The biased encoder takes the past trajectory $x$, a risk-level $\sigma$, and the robot future trajectory $y_\text{robot}$. It outputs the parameters of a Normal distribution $\mu^{(b)}$ and $\log(\operatorname{diag}(\Sigma^{(b)}))$.
    \item The decoder takes the past trajectory $x$ and a latent sample $z$ and outputs a prediction $y$.
\end{itemize}

 The time-sequence trajectories are flattened before fed into the model. Conversely, the output of the model is reshaped back into time-sequence trajectories.
 Each MLP is composed of 3 fully connected layers with a hidden dimension of 64 and ReLU activations. We chose a latent space dimension of 2 that is enough to show a working model and allows us to represent the latent space in 2D plots.
The overall model is defined with 54.4K parameters.

\begin{figure}[!h]
    \centering
    \includegraphics[trim=1.5cm 0.5cm 0.5cm 1.5cm, clip, width=0.6\columnwidth]{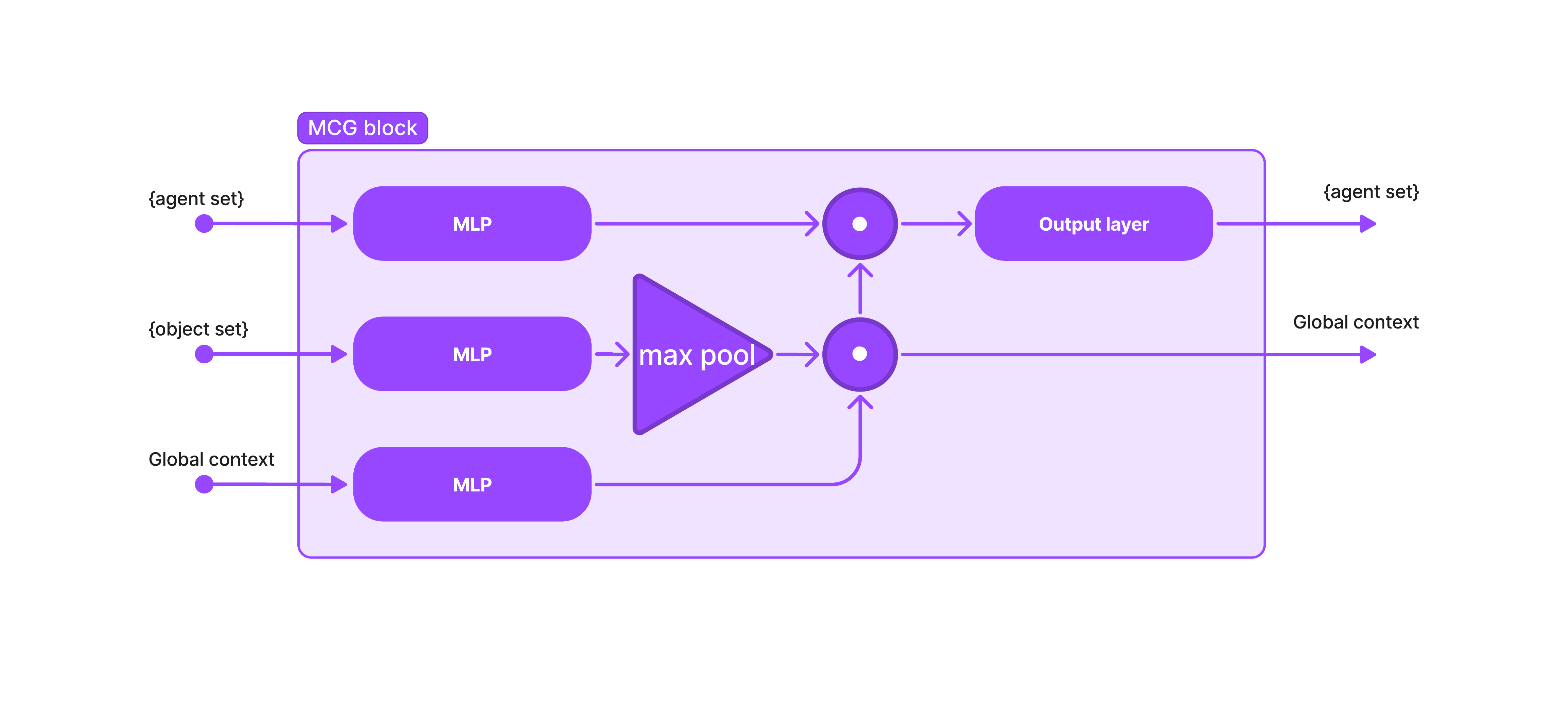}
    \caption{Diagram of the modified CG blocks (original CG blocks are defined in~\cite{varadarajan2021multipath++}). The circle dot is an element-wise multiplication (with each vector of the set).}
    \label{fig:modified_cg}
\end{figure}

Our second architecture resembles state-of-the-art forecasting architectures used in challenging, real-world scenarios. It is designed to model the interactions of the agent to be predicted with the surrounding agents and map elements. It also takes the form of a CVAE architecture and uses two MLP encoders and an MLP decoder similar to those described above, but with additional context inputs and larger hidden dimensions.
We chose a latent space dimension of 16 because that gave satisfactory results in terms of final displacement error.
The social and map interactions are accounted for with a modified multi-context gating block \cite{varadarajan2021multipath++} composed of 3 context gating blocks.
The context gating blocks each count three MLP modules with a hidden dimension of 256 (twice the input dimension).
The MLP modules each count three layers and ReLU activations.
Our modified context-gating block is represented \cref{fig:modified_cg}.
We stack these modified CG blocks with a running average of their output exactly as in~\cite{varadarajan2021multipath++}.

\begin{figure}[!h]
    \centering
    \includegraphics[trim=1.5cm 0.5cm 0.5cm 1.5cm, clip, width=\columnwidth]{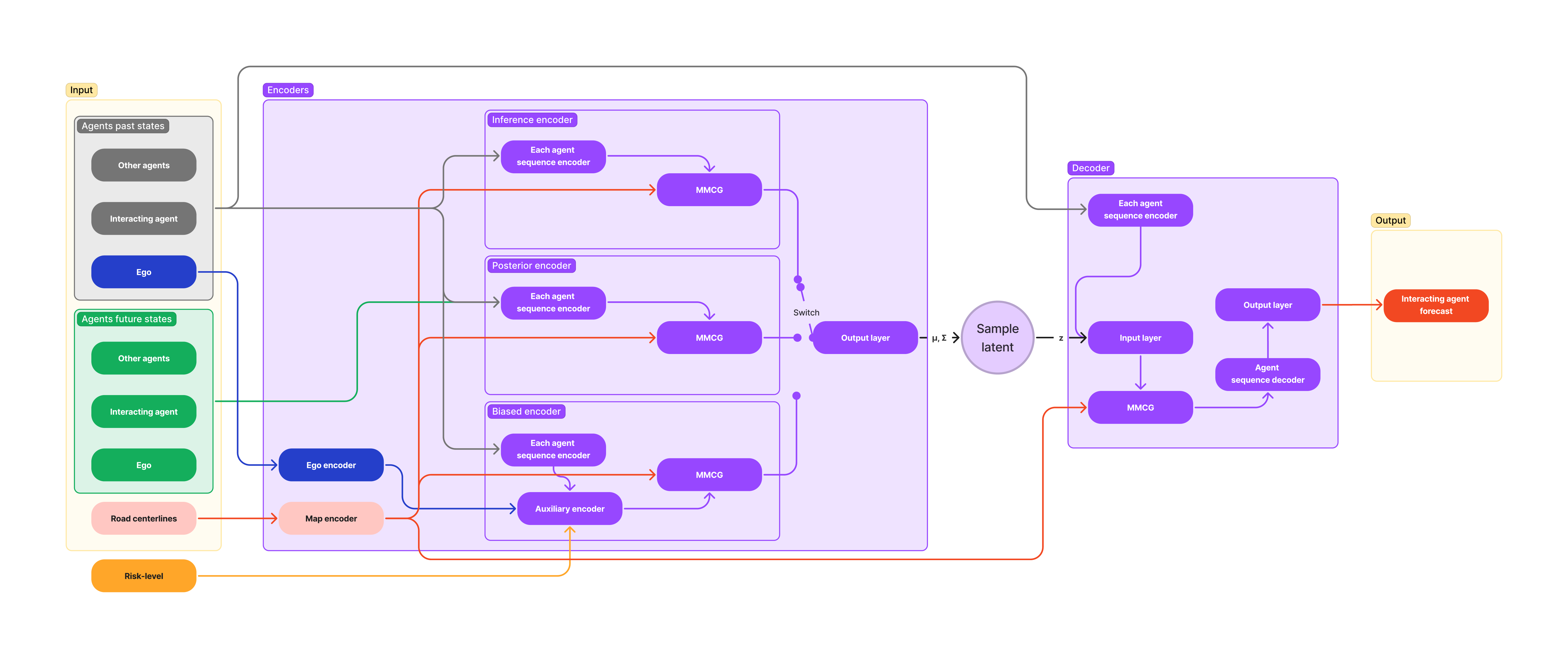}
    \caption{Diagram of the model used for the WOMD experiment. MMCG stands for modified multi-context gating blocks. MCG blocks are defined in~\cite{varadarajan2021multipath++}.}
    \label{fig:waymo_pred}
\end{figure}

The overall model represented in \cref{fig:waymo_pred} counts 15.8M parameters.

\subsection{Model complexity}

The CVAE model was trained for 6 hours 20 minutes on a single GPU Nvidia Titan Xp. Then its parameters are frozen and the biased encoder was trained for 4 days and 10 hours on the same GPU. This second training is time consuming because it involves the estimation of the risk with 64 then 256 samples which multiplies the tensor dimension and requires a smaller batch size to be used to fit in the GPU memory. This is exactly the computation overhead that our proposed method reduces at inference by reducing the number of samples needed for the risk estimation.

Because we only use fully connected layers, the overall complexity of the model is $O(b\times s\times a \times t\times f \times h) + O(b \times s \times a \times h^2) + O(b \times o \times m_s\times m_f\times h)$ with batch size $b$, number of samples $s$, number of agents $a$, hidden feature dimension $h$, time sequence length $t$, input feature dimension $f$, map element sequence length $m_s$, map input feature dimension $m_f$. 
With our choices of hyperparameters $s\times a \times t\times f>o \times m_s \times m_f$ and $t\times f>h$ so the complexity is $O(b\times s\times a \times (t\times f) \times h)$. At inference time, the number of samples $s$ can be kept small using our method and the batch dimension is 1. The most expensive operation is the first matrix multiplication but this operation can be easily parallelized and is often well optimized. The most limiting aspect might be the memory footprint. At test time with 20 samples, the allocated GPU memory reaches almost 2GiB.

\section{Additional Experimental Details and Results}
\label{sec: experiment_details}
\subsection{Simulation Experiments}
\subsubsection{Biasing forecasts}
\cref{tab:sim-results} shows the distance metrics and risk error of the prediction model in our simulated environment (\cref{fig: road_scene}). In this didactic environment, the model is performing well with low distance-based errors and a low risk error.

\begin{table*}[!h]
\centering 
\vspace{10pt}
\caption{Motion forecasting error and risk estimation error on the simulation validation set. 
\textbf{minFDE (K)}: minimum final displacement error over K samples,
\textbf{risk error (K)}: mean value of the signed difference between the average cost of the biased forecasts over K samples and the risk estimation using the unbiased forecasts,
\textbf{risk $|$error$|$ (K)}: mean value of the absolute values of the risk estimation error.}  
\label{tab:sim-results}
\resizebox{0.9\textwidth}{!}{
\begin{tabular}{lccccc}
\hline
Predictive Model & $\sigma$ & minFDE (16) & FDE (1) & Risk error (4) & Risk $|$error$|$ (4) \\
\hline
Unbiased CVAE & \scriptsize{NA} &$0.45$ \scriptsize{$\pm 0.00$}& $0.81$ \scriptsize{$\pm 0.00$}& \scriptsize{NA} & \scriptsize{NA} \\
\hline
Biased CVAE (\modelacronym) & 0 & $0.48$ \scriptsize{$\pm 0.00$}& $0.80$ \scriptsize{$\pm 0.00$}& $0.00$ \scriptsize{$\pm 0.00$}& $0.01$ \scriptsize{$\pm 0.00$}\\
Biased CVAE (\modelacronym) & 0.3 & $0.51$ \scriptsize{$\pm 0.00$}& $0.80$ \scriptsize{$\pm 0.00$}& $0.00$ \scriptsize{$\pm 0.00$}& $0.01$ \scriptsize{$\pm 0.00$}\\
Biased CVAE (\modelacronym) & 0.5 & $0.54$ \scriptsize{$\pm 0.00$}& $0.82$ \scriptsize{$\pm 0.00$}& $0.00$ \scriptsize{$\pm 0.00$}& $0.01$ \scriptsize{$\pm 0.00$}\\
Biased CVAE (\modelacronym) & 0.8 & $0.62$ \scriptsize{$\pm 0.01$}& $0.91$ \scriptsize{$\pm 0.00$}& $0.00$ \scriptsize{$\pm 0.00$}& $0.01$ \scriptsize{$\pm 0.00$}\\
Biased CVAE (\modelacronym) & 0.95 & $0.74$ \scriptsize{$\pm 0.01$}& $1.05$ \scriptsize{$\pm 0.00$}& $0.01$ \scriptsize{$\pm 0.00$}& $0.01$ \scriptsize{$\pm 0.00$}\\
Biased CVAE (\modelacronym) & 1 & $0.81$ \scriptsize{$\pm 0.01$}& $1.11$ \scriptsize{$\pm 0.00$}& $-0.02$ \scriptsize{$\pm 0.00$}& $0.03$ \scriptsize{$\pm 0.00$}\\
\hline
\end{tabular}
 }
\end{table*}

\begin{figure}[!h]
    \vspace{-10pt}
    \centering
    \begin{subfigure}[b]{0.32\columnwidth}
        \centering
        \adjustbox{width=\columnwidth, trim=0cm 0cm 0cm 0.9cm, clip}{%
            \includegraphics{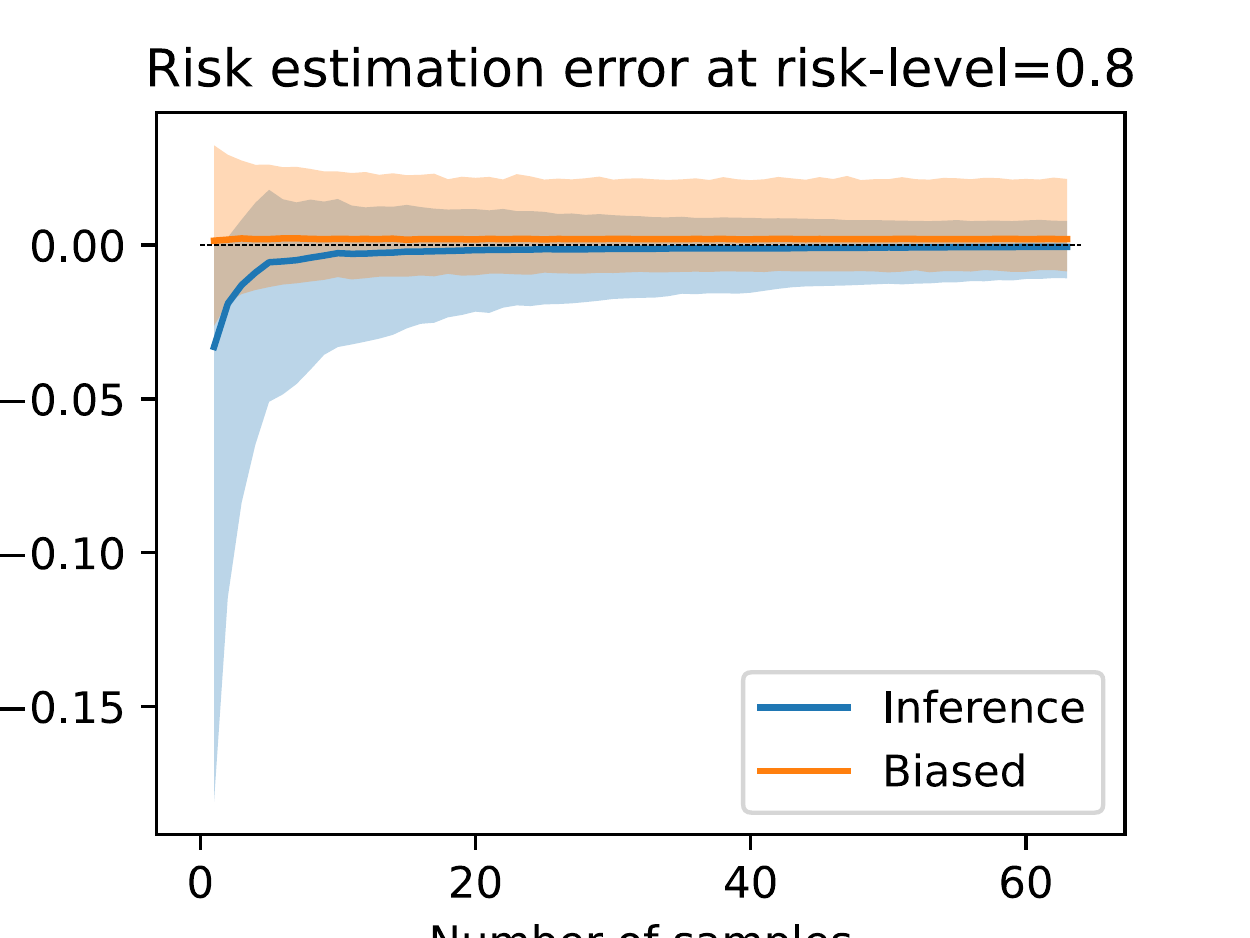}
        }
        \caption{Risk-level $\sigma=0.8$}
        \label{subfig: risk_samples_08}
    \end{subfigure}
    \hfill
    \begin{subfigure}[b]{0.32\columnwidth}
        \centering
        \adjustbox{width=\columnwidth, trim=0cm 0cm 0cm 0.9cm, clip}{%
            \includegraphics{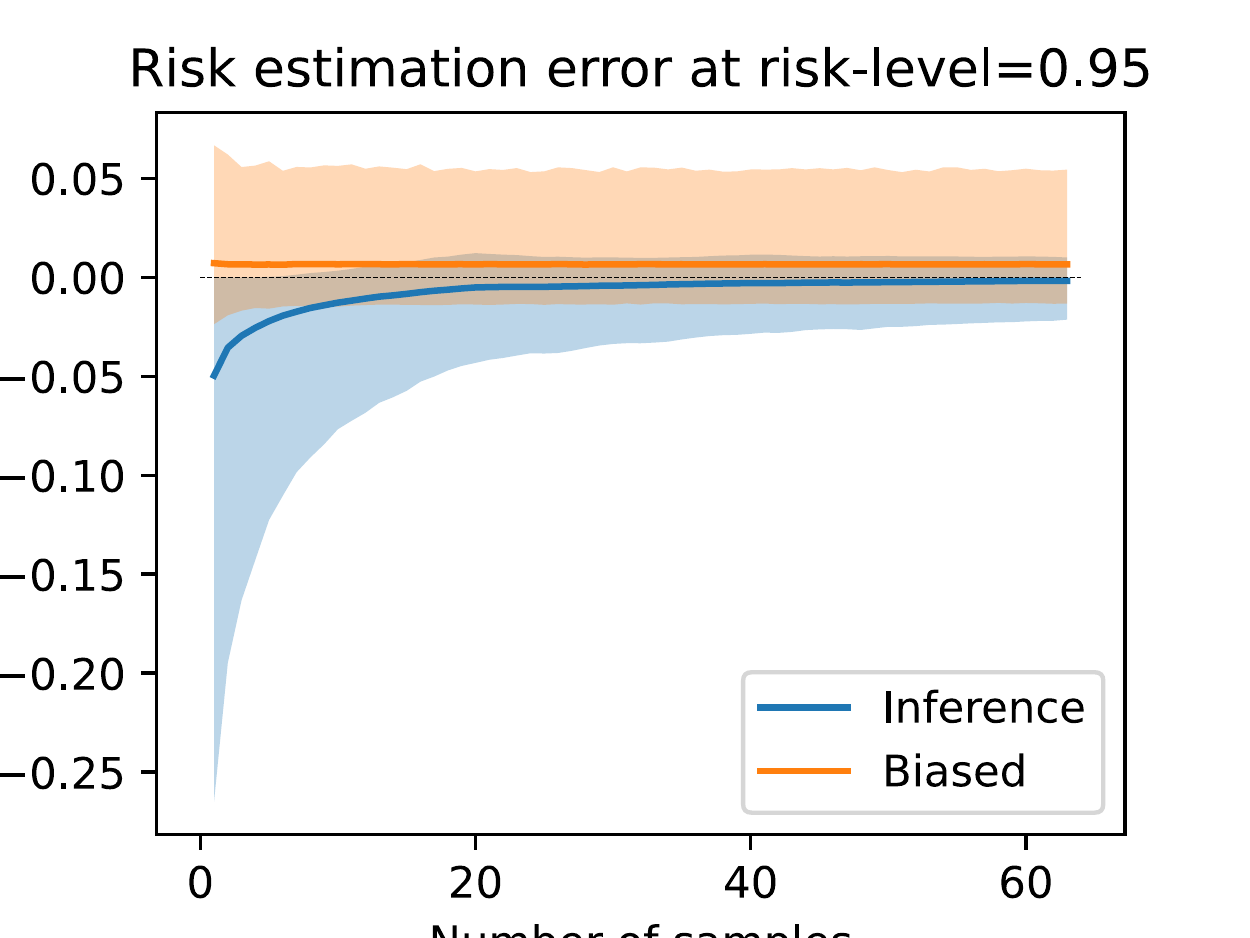}
        }        \caption{Risk-level $\sigma=0.95$}
        \label{subfig: risk_samples_095}
    \end{subfigure}
    \hfill
    \begin{subfigure}[b]{0.32\columnwidth}
        \centering
        \adjustbox{width=\columnwidth, trim=0cm 0cm 0cm 0.9cm, clip}{%
            \includegraphics{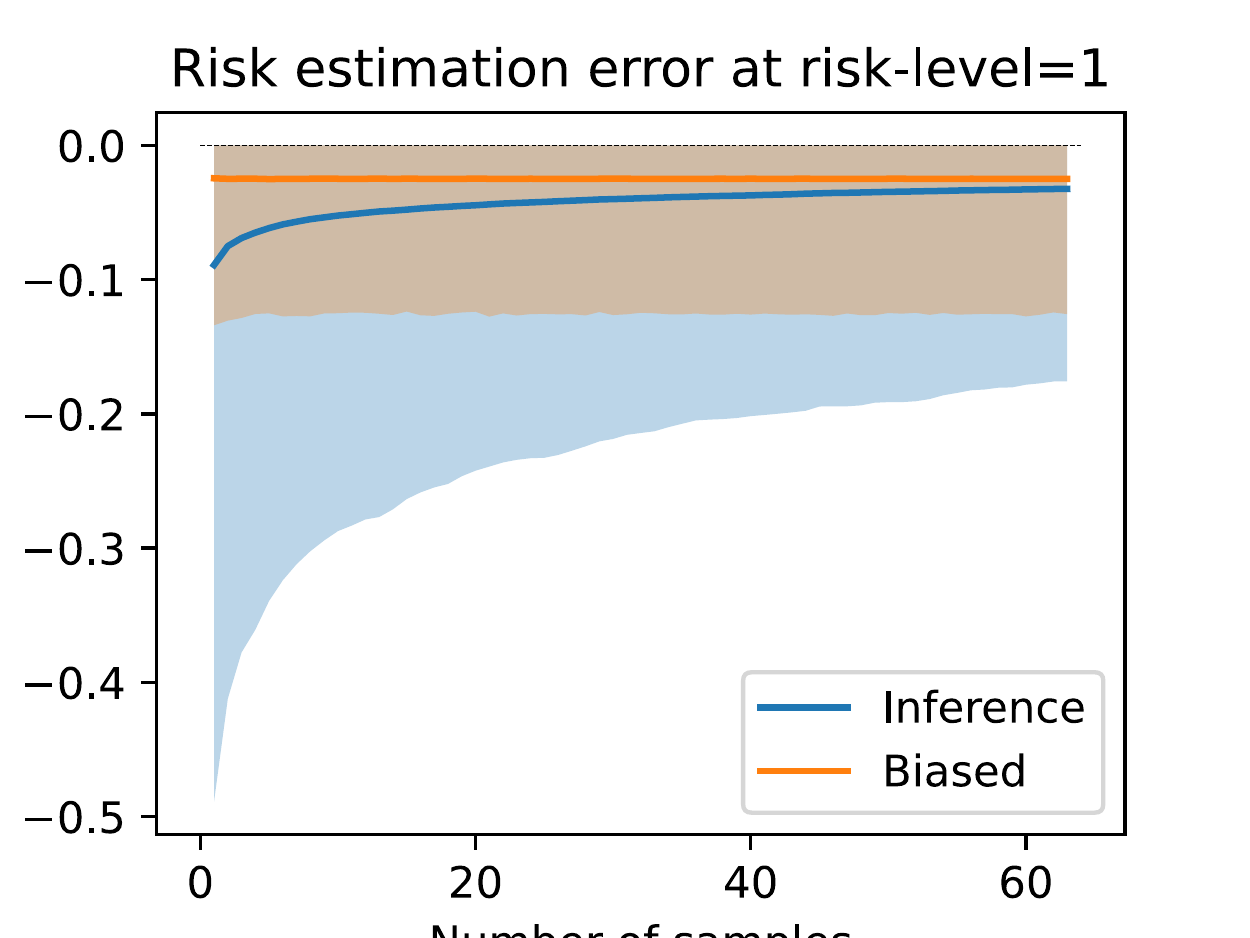}
        }        \caption{Risk-level $\sigma=1$}
        \label{subfig: risk_samples_1}
    \end{subfigure}
    \hfill
    \vspace{10pt}
    \caption{Risk estimation error using the mean cost of the biased forecasts (biased) or the Monte-Carlo estimate of the risk under the unbiased forecasts (inference) as functions of the number of samples at different risk levels. Shaded regions indicate the 5\% and 95\% quantiles. The ``ground-truth" used as a reference is computed with a Monte-Carlo estimate of the risk under the unbiased forecast using 4096 samples.}
    \label{fig:risk_samples_sim}
\end{figure}

\cref{fig:risk_samples_sim} compares the Monte-Carlo risk estimation under the unbiased prediction (noted inference) and our proposed approach for risk estimation using biased predictions (noted biased).
The plots show the average risk estimation error over the validation set as functions of the number of samples.
The reference risk is computed with a Monte-Carlo risk estimation using 4096 samples.
These plots show that our method may over-estimate risk more often (the 95\% quantile is higher for our method across all number of samples). It also shows that our method makes an almost unbiased risk estimation even with only one sample while the Monte-Carlo approach underestimates the risk more often (the 5\% quantile is lower than our method especially at low numbers of samples).

\subsubsection{Planning with a biased prediction}
\label{sec:experimental_details_mpc}
\begin{figure}[!h]
    \centering
    \adjustbox{width=0.4\columnwidth, trim=0 0 0 0, clip}{%
        \includegraphics{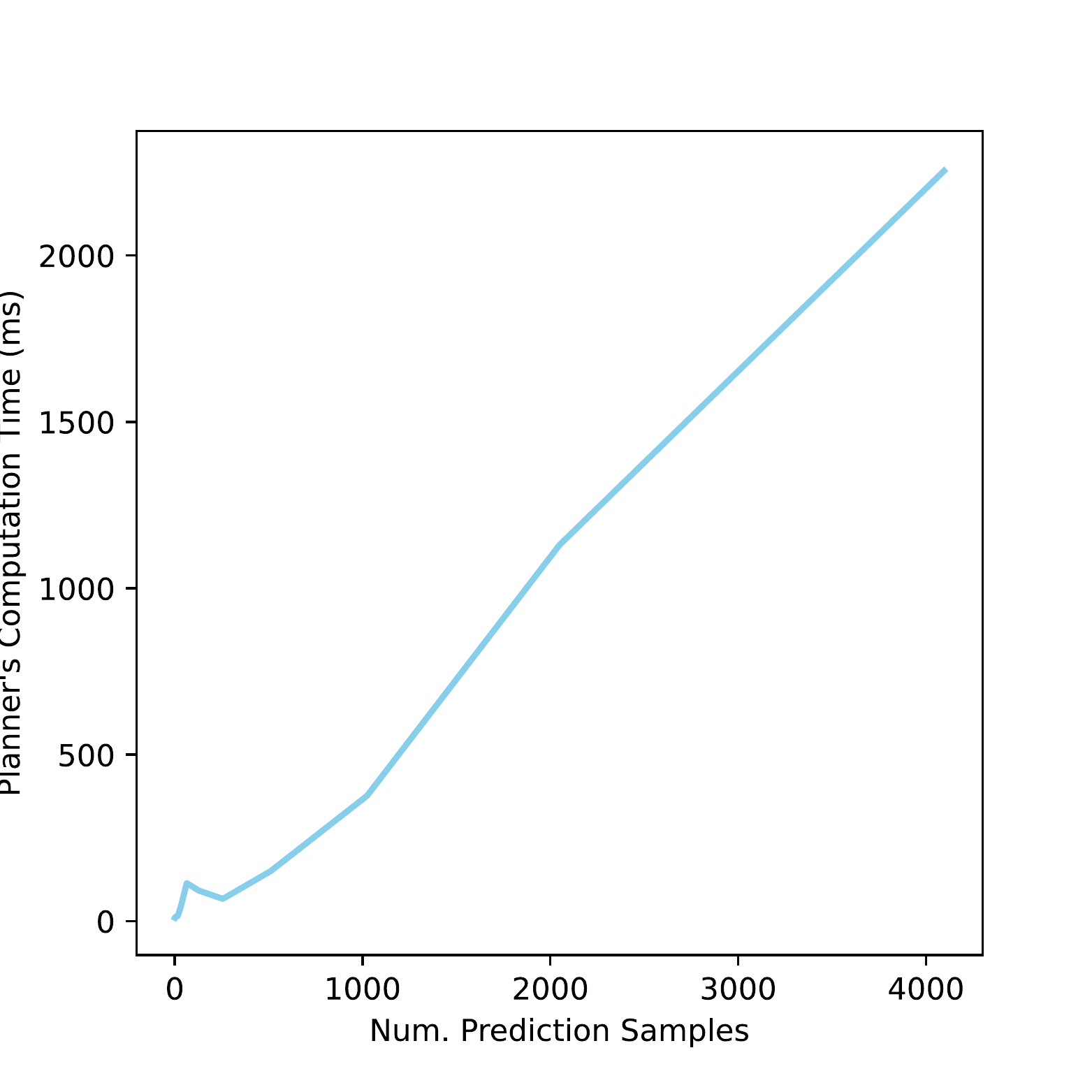}
    }
    \caption{
            Overall computation time required by the planner at different values of $n_{\text{samples}}$.
    }
    \label{fig:planner_computation_time}
\end{figure}

We provide additional details and results on the CEM planning experiment presented in Section \ref{sec: didactic_biasing_with_mpc}.
Starting from $y_{\text{robot}}^{\text{init}}$, CEM iteratively updates the planned robot trajectory $y_{\text{robot}}$ by first sampling $n^{\text{robot}}_{\text{samples}} = 100$ robot trajectories in each iteration, from a Gaussian distribution with the mean being the $y_{\text{robot}}$ of the previous iteration.
For each trajectory, we evaluate the following objective and select $n^{\text{robot}}_{\text{elites}} = 30$ elite trajectories that achieve the lowest objective values:
\begin{align}
    \mathcal{L}_{\text{plan}}(y_{\text{robot}}) = \mathcal{R}_p(J^{y_{\text{robot}}}(Y), \sigma) + \Vert y_{\text{robot}} - y_{\text{ref}} \Vert_{Q}^2,
    \label{eq:planner_objective}
\end{align}
wherein the first term is the risk measure of the TTC cost (see Section \ref{sec: ttc_cost}) with respect to the stochastic human future trajectory $Y$,
and the second term is the quadratic tracking cost with respect to a given reference trajectory $y_{\text{ref}}$ under a symmetric, positive semi-definite cost matrix $Q$.
The risk is estimated using $n_{\text{samples}}$ prediction samples of $Y$, through the Monte Carlo CVaR estimator~\cite{hong2014monte} (for the unbiased CVAE) or Monte Carlo expectation (for the proposed RAP).
The reference trajectory $y_{\text{ref}}$ is chosen to be a constant velocity trajectory at the desired speed of 14 m/s.
Once the elites are chosen, CEM updates $y_{\text{robot}}$ to be the average of the elites.
This iteration is repeated $n_{\text{iter}} = 10$ times, resulting in the overall time complexity of $O(n_{\text{iter}} \times n_{\text{samples}}^{\text{robots}} \times n_{\text{samples}})$.
In particular it is linear in $n_{\text{samples}}$.
This is confirmed in \cref{fig:planner_computation_time}, when the planner was run on an Intel(R) Xeon(R) W-3345 CPU @ 3.00GHz.
Leveraging a GPU would further reduce the overall run-time, but would also come at the cost of GPU memory consumption and may hinder other modules (such as perception) from performing real-time information processing.
We refer the reader to prior work~\cite{chua2018deep, nagabandi2020deep} for further details of CEM.

\begin{figure}[h]
    \centering
    \adjustbox{width=1.0\columnwidth, trim=2cm 0cm 2cm 0cm, clip}{%
    \includegraphics{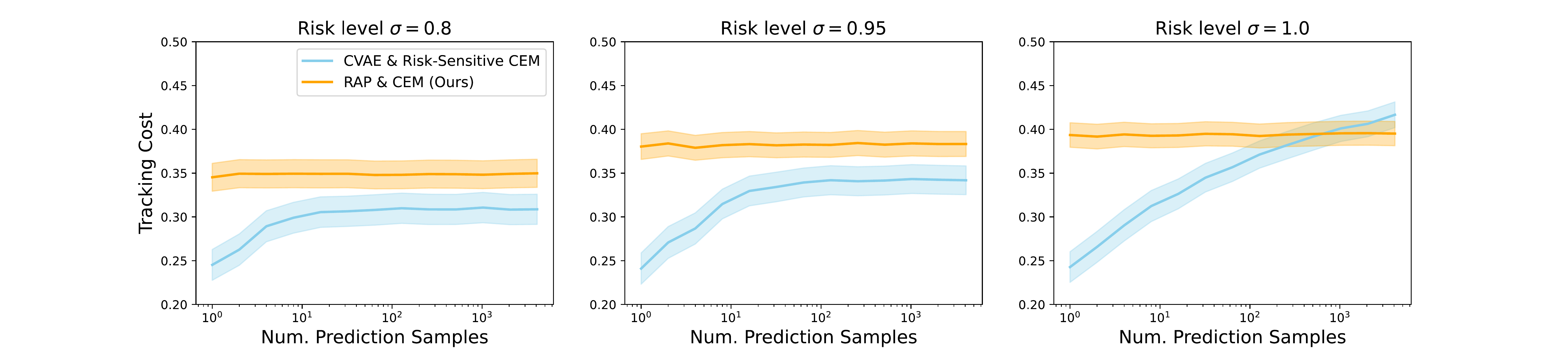}
    }
    \caption{Tracking cost of the optimized $y_{\text{robot}}^*$, averaged over 500 episodes (lower the better). Ribbons show 95\% confidence intervals of the mean. }
    \label{fig:stats_mpc_tracking}
\end{figure}

In order to visualize the interplay of the two terms in the objective \eqref{eq:planner_objective}, we present in \cref{fig:stats_mpc_tracking} the tracking cost plots from the same experiment as in Section \ref{sec: didactic_biasing_with_mpc}.
The only difference between \cref{fig: stats_mpc} and \cref{fig:stats_mpc_tracking} is that the y-axis of \cref{fig: stats_mpc} shows the ground-truth TTC cost $J^{y_{\text{robot}}^*}(y)$ whereas that of \cref{fig:stats_mpc_tracking} measures the tracking cost $\Vert y_{\text{robot}}^* - y_{\text{ref}} \Vert_{Q}^2$.
The results depicted in \cref{fig:stats_mpc_tracking} supports our claim in Section \ref{sec: didactic_biasing_with_mpc}, which is that the planner over-optimistically optimizes trajectory tracking to the detriment of safety for the baseline CVAE predictor.
Indeed, with fewer than 16 samples the baseline tends to achieve noticeably lower tracking costs at all risk-sensitivity levels.
This is because the risk term in equation \eqref{eq:planner_objective} is under-estimated due to the limited number of prediction samples. In turn, the optimization overestimates the relative importance of the tracking cost term.
The proposed RAP predictor results in slightly higher tracking costs than the baseline at $\sigma = 0.8$ and $0.95$, but its performance is not affected by the number of prediction samples $n_{\text{samples}}$ that we draw from the risk-aware predictor for robust planning.

\begin{figure}[!h]
    \begin{subfigure}[b]{1.0\columnwidth}
        \centering
        \adjustbox{width=1.0\columnwidth}{%
            \includegraphics{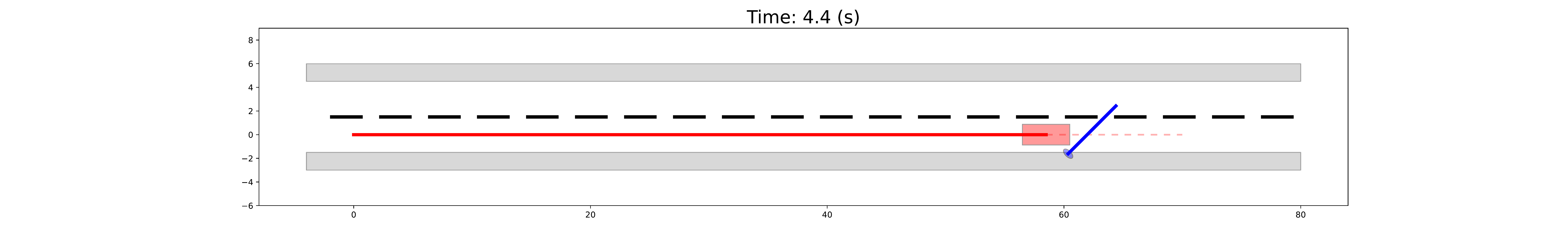}
        }
        \caption{
                CVAE \& Risk-Sensitive CEM
        }
        \label{fig:robot_trajectory_baseline}
    \end{subfigure}
    
    \hfill
    
    \begin{subfigure}[b]{1.0\columnwidth}
        \centering
        \adjustbox{width=1.0\columnwidth}{%
            \includegraphics{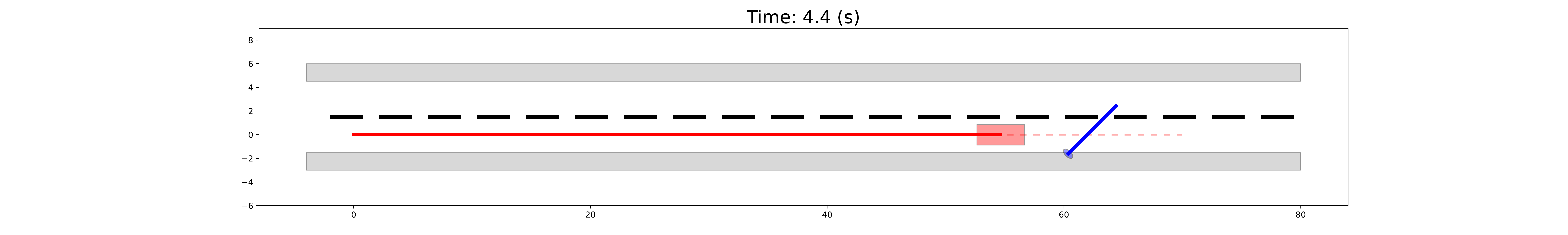}
        }
        \caption{
                RAP \& CEM (Ours)
        }
        \label{fig:robot_trajectory_rap}
    \end{subfigure}
    
    \hfill
    
    \caption{
    Optimized robot trajectory $y^*_{\text{robot}}$, executed and paused at $t = 4.4$ (s) in a test episode. Solid colored lines represent the executed trajectories of the robot and the pedestrian, and the dashed red line the robot's reference trajectory. (a) The robot almost collided with a pedestrian when the risk was taken into account in the planner. (b) When the risk was taken into account in the predictor, the robot slowed down slightly to keep more distance from the pedestrian. In both (a) and (b), the risk-sensitivity level was $\sigma = 0.95$ and $n_{\text{samples}} = 1$ prediction sample was given to the planner.
    }
    \label{fig:robot_trajectory_comparison}
\end{figure}

Lastly, \cref{fig:robot_trajectory_comparison} illustrates the difference in the optimized robot trajectory $y^*_{\text{robot}}$ between the baseline and the proposed RAP.
The two trajectories of the robot provide a qualitative explanation to the statistical results presented in \cref{fig: stats_mpc} and \cref{fig:stats_mpc_tracking}.
That is, our proposed framework appropriately slowed down the robot to keep more distance from the pedestrian, yielding a lower TTC cost and a higher tracking cost compared to the conventional risk-sensitive prediction-planning approach.

\subsection{Experiments on real-world data}
\begin{figure}[!h]
    \vspace{-10pt}
    \centering
    \begin{subfigure}[b]{0.32\columnwidth}
        \centering
        \adjustbox{width=\columnwidth, trim=0cm 0cm 0cm 0.9cm, clip}{%
            \includegraphics{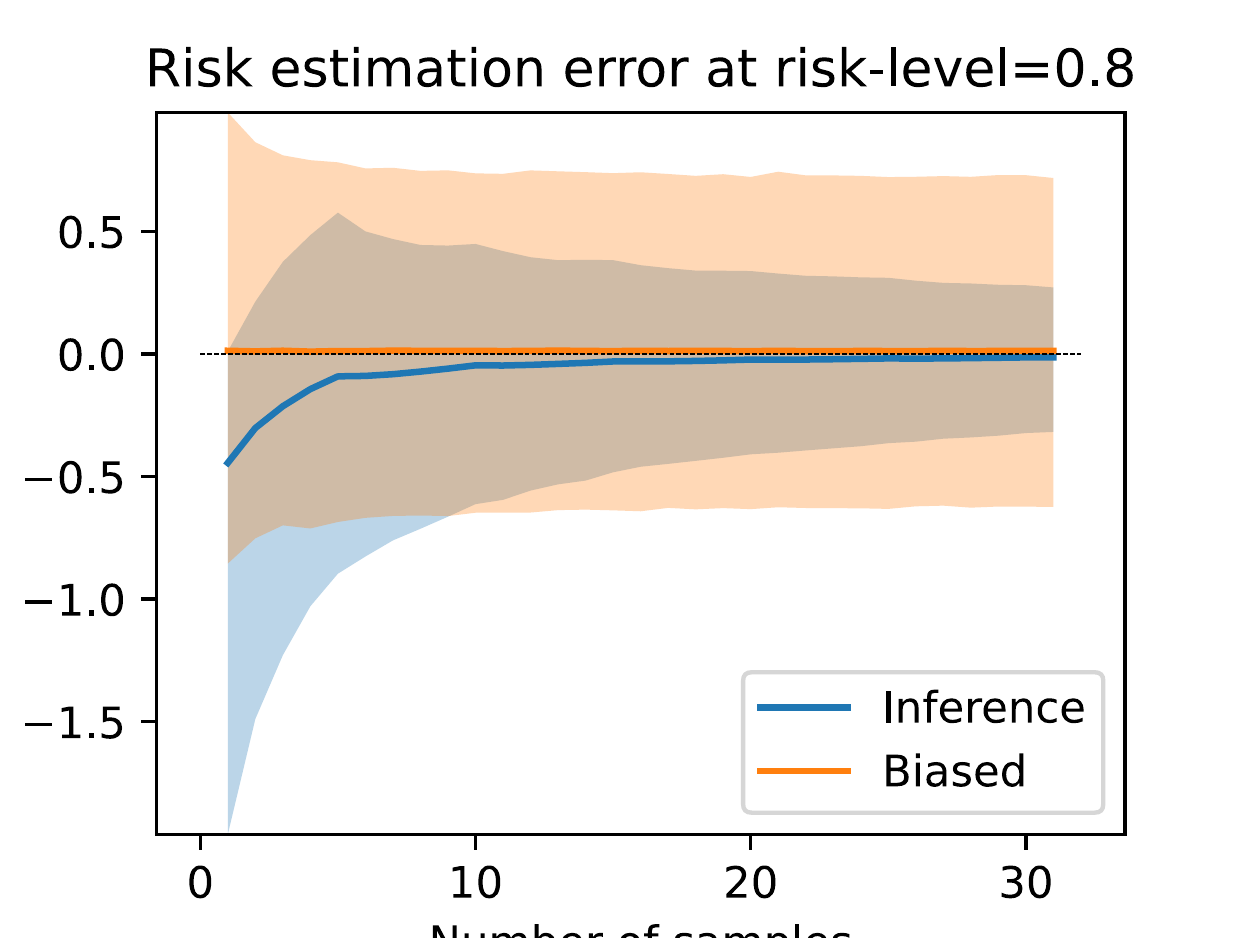}
        }
        \caption{Risk-level $\sigma=0.8$}
        \label{subfig: risk_samples_08_2}
    \end{subfigure}
    \hfill
    \begin{subfigure}[b]{0.32\columnwidth}
        \centering
        \adjustbox{width=\columnwidth, trim=0cm 0cm 0cm 0.9cm, clip}{%
            \includegraphics{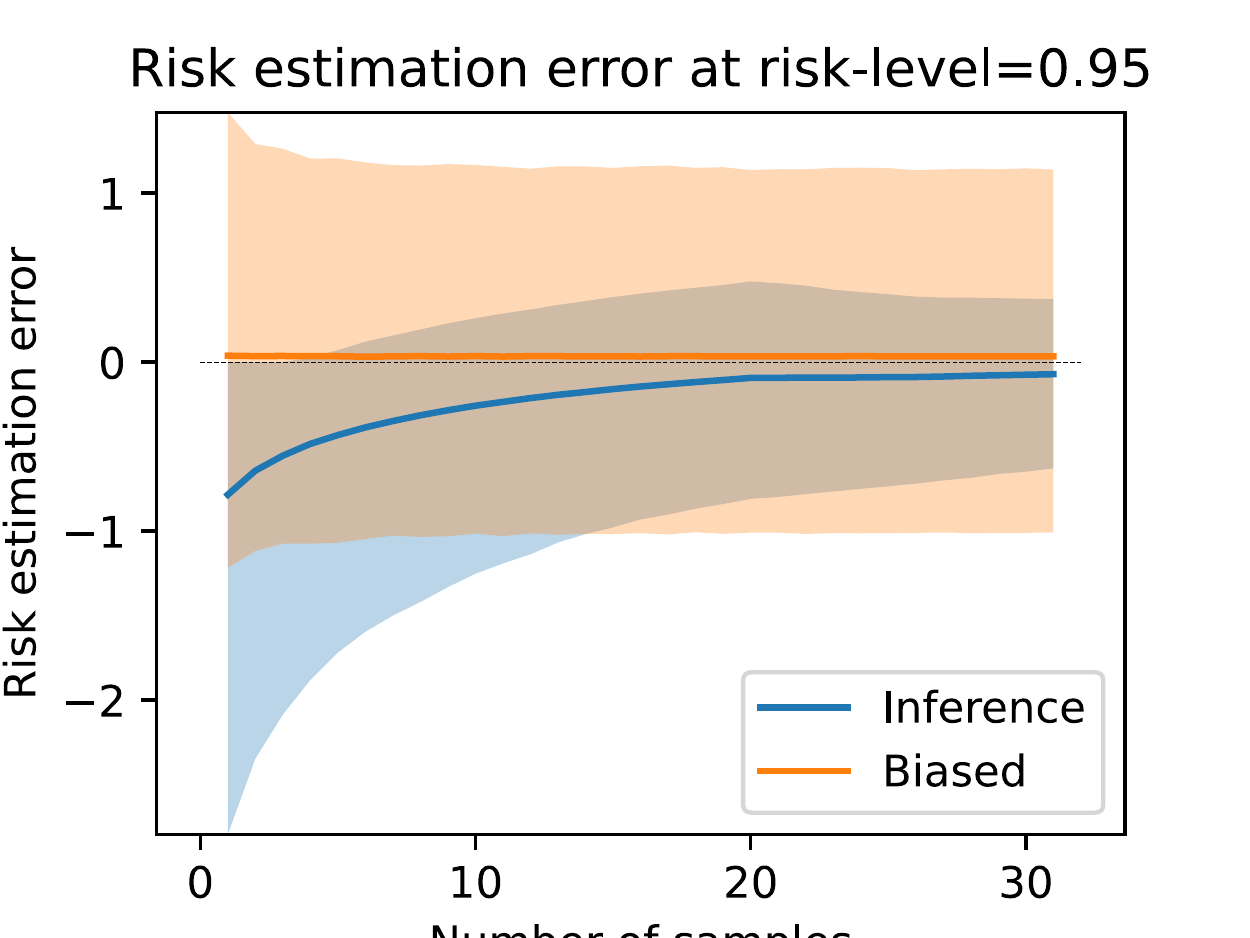}
        }        \caption{Risk-level $\sigma=0.95$}
        \label{subfig: risk_samples_095_2}
    \end{subfigure}
    \hfill
    \begin{subfigure}[b]{0.32\columnwidth}
        \centering
        \adjustbox{width=\columnwidth, trim=0cm 0cm 0cm 0.9cm, clip}{%
            \includegraphics{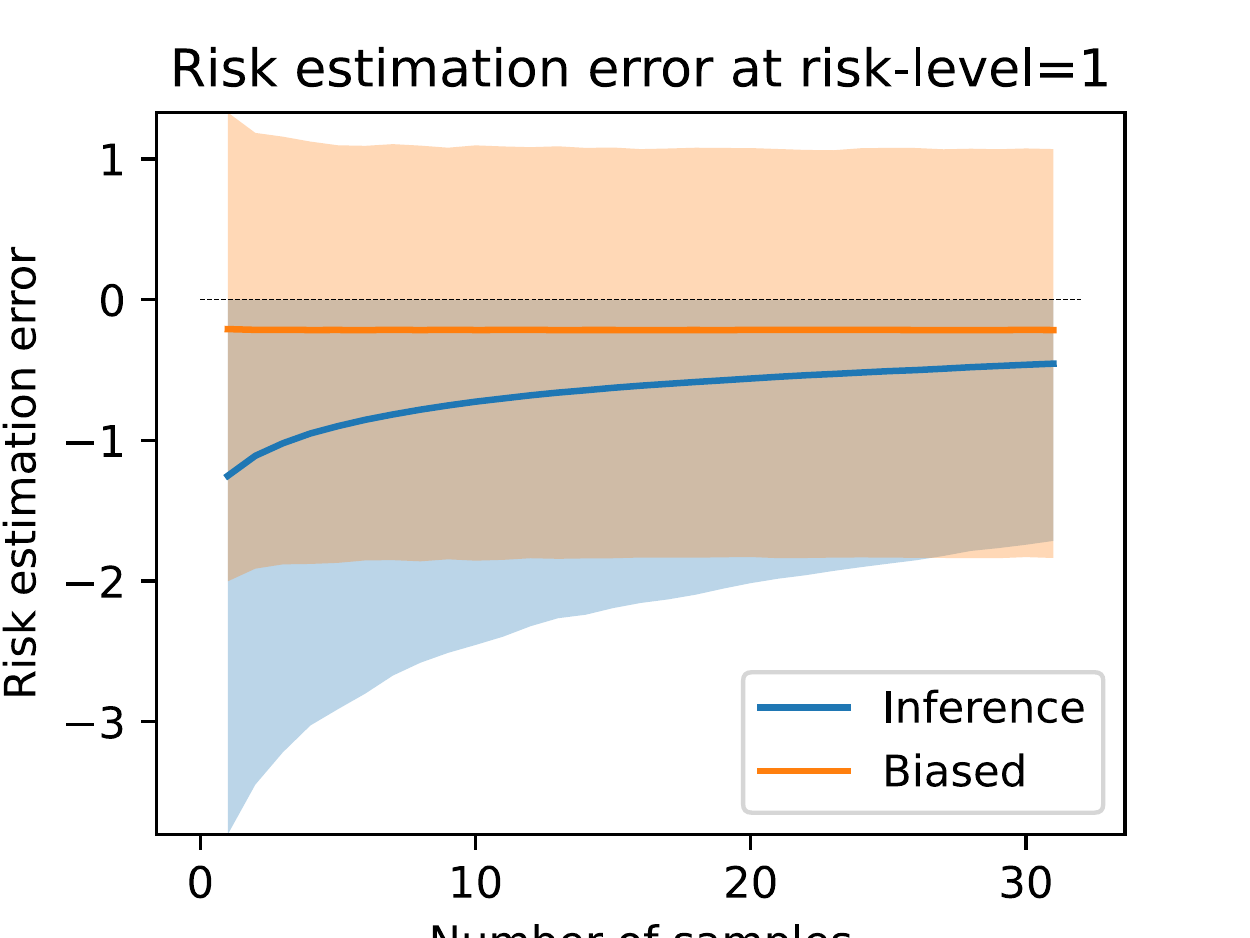}
        }        \caption{Risk-level $\sigma=1$}
        \label{subfig: risk_samples_1_2}
    \end{subfigure}
    \hfill
    \vspace{10pt}
    \caption{Risk estimation error using the mean cost of the biased forecasts (biased) or the Monte-Carlo estimate of the risk under the unbiased forecasts (inference) as functions of the number of samples at different risk levels. Shaded regions indicate the 5\% and 95\% quantiles.}
    \label{fig:risk_samples}
\end{figure}

\cref{fig:risk_samples} compares the error from two methods for risk estimation as functions of the number of samples. Monte-Carlo risk estimation using the inference forecast distribution systematically underestimates the risk, especially with few samples. Our proposed method shows little estimation bias when the risk-level is below $0.95$, and this relatively-small amount of bias is independent of the number of samples.

\begin{figure}
\foreach \i/\cap in {%
    scene_102_risk_0.95_n_samples_16_use_biaser_True.pdf/Biased,
    scene_102_risk_0_n_samples_16_use_biaser_True.pdf/Unbiased,
    scene_106_risk_0.95_n_samples_16_use_biaser_True.pdf/Biased,
    scene_106_risk_0_n_samples_16_use_biaser_True.pdf/Unbiased,
    scene_107_risk_0.95_n_samples_16_use_biaser_True.pdf/Biased,
    scene_107_risk_0_n_samples_16_use_biaser_True.pdf/Unbiased,
    scene_108_risk_0.95_n_samples_16_use_biaser_True.pdf/Biased,
    scene_108_risk_0_n_samples_16_use_biaser_True.pdf/Unbiased,
    scene_109_risk_0.95_n_samples_16_use_biaser_True.pdf/Biased,
    scene_109_risk_0_n_samples_16_use_biaser_True.pdf/Unbiased,
    scene_110_risk_0.95_n_samples_16_use_biaser_True.pdf/Biased,
    scene_110_risk_0_n_samples_16_use_biaser_True.pdf/Unbiased,
    scene_111_risk_0.95_n_samples_16_use_biaser_True.pdf/Biased,
    scene_111_risk_0_n_samples_16_use_biaser_True.pdf/Unbiased,
    scene_113_risk_0.95_n_samples_16_use_biaser_True.pdf/Biased,
    scene_113_risk_0_n_samples_16_use_biaser_True.pdf/Unbiased,
    scene_116_risk_0.95_n_samples_16_use_biaser_True.pdf/Biased,
    scene_116_risk_0_n_samples_16_use_biaser_True.pdf/Unbiased,
    scene_117_risk_0.95_n_samples_16_use_biaser_True.pdf/Biased,
    scene_117_risk_0_n_samples_16_use_biaser_True.pdf/Unbiased,
    scene_118_risk_0.95_n_samples_16_use_biaser_True.pdf/Biased,
    scene_118_risk_0_n_samples_16_use_biaser_True.pdf/Unbiased,
    scene_11_risk_0.95_n_samples_16_use_biaser_True.pdf/Biased,
    scene_11_risk_0_n_samples_16_use_biaser_True.pdf/Unbiased,
    scene_120_risk_0.95_n_samples_16_use_biaser_True.pdf/Biased,
    scene_120_risk_0_n_samples_16_use_biaser_True.pdf/Unbiased,
    scene_12_risk_0.95_n_samples_16_use_biaser_True.pdf/Biased,
    scene_12_risk_0_n_samples_16_use_biaser_.pdf/Unbiased,
    scene_16_risk_0.95_n_samples_16_use_biaser_True.pdf/Biad,
    scene_16_risk_0_n_samples_16_use_biaser_True.pdf/Unbiased,
    scene_200_risk_0.95_n_samples_16_use_biaser_True.pdf/Biased,
    scene_200_risk_0_n_samples_16_use_biaser_True.pdf/Unbiased,
    scene_201_risk_0.95_n_samples_16_use_biaser_True.pdf/Biased,
    scene_201_risk_0_n_samples_16_use_biaser_True.pdf/Unbiased,
    scene_206_risk_0.95_n_samples_16_use_biaser_True.pdf/Biased,
    scene_206_risk_0_n_samples_16_use_biaser_True.pdf/Unbiased,
    scene_210_risk_0.95_n_samples_16_use_biaser_True.pdf/Biased,
    scene_210_risk_0_n_samples_16_use_biaser_True.pdf/Unbiased,
    scene_212_risk_0.95_n_samples_16_use_biaser_True.pdf/Biased,
    scene_212_risk_0_n_samples_16_use_biaser_True.pdf/Unbiased,
    scene_21_risk_0.95_n_samples_16_use_biaser_True.pdf/Biased,
    scene_21_risk_0_n_samples_16_use_biaser_True.pdf/Unbiased,
    scene_22_risk_0.95_n_samples_16_use_biaser_True.pdf/Biased,
    scene_22_risk_0_n_samples_16_use_biaser_True.pdf/Unbiased,
    scene_7_risk_0.95_n_samples_16_use_biaser_True.pdf/Biased,
    scene_7_risk_0_n_samples_16_use_biaser_True.pdf/Unbiased,
    scene_299_risk_0.95_n_samples_16_use_biaser_True.pdf/Biased,
    scene_299_risk_0_n_samples_16_use_biaser_True.pdf/Unbiased
} {%
    \begin{subfigure}[b]{0.14\columnwidth}
        \adjustbox{width=\columnwidth, trim=2cm 1cm 3.5cm 2cm, clip}{%
            \includegraphics{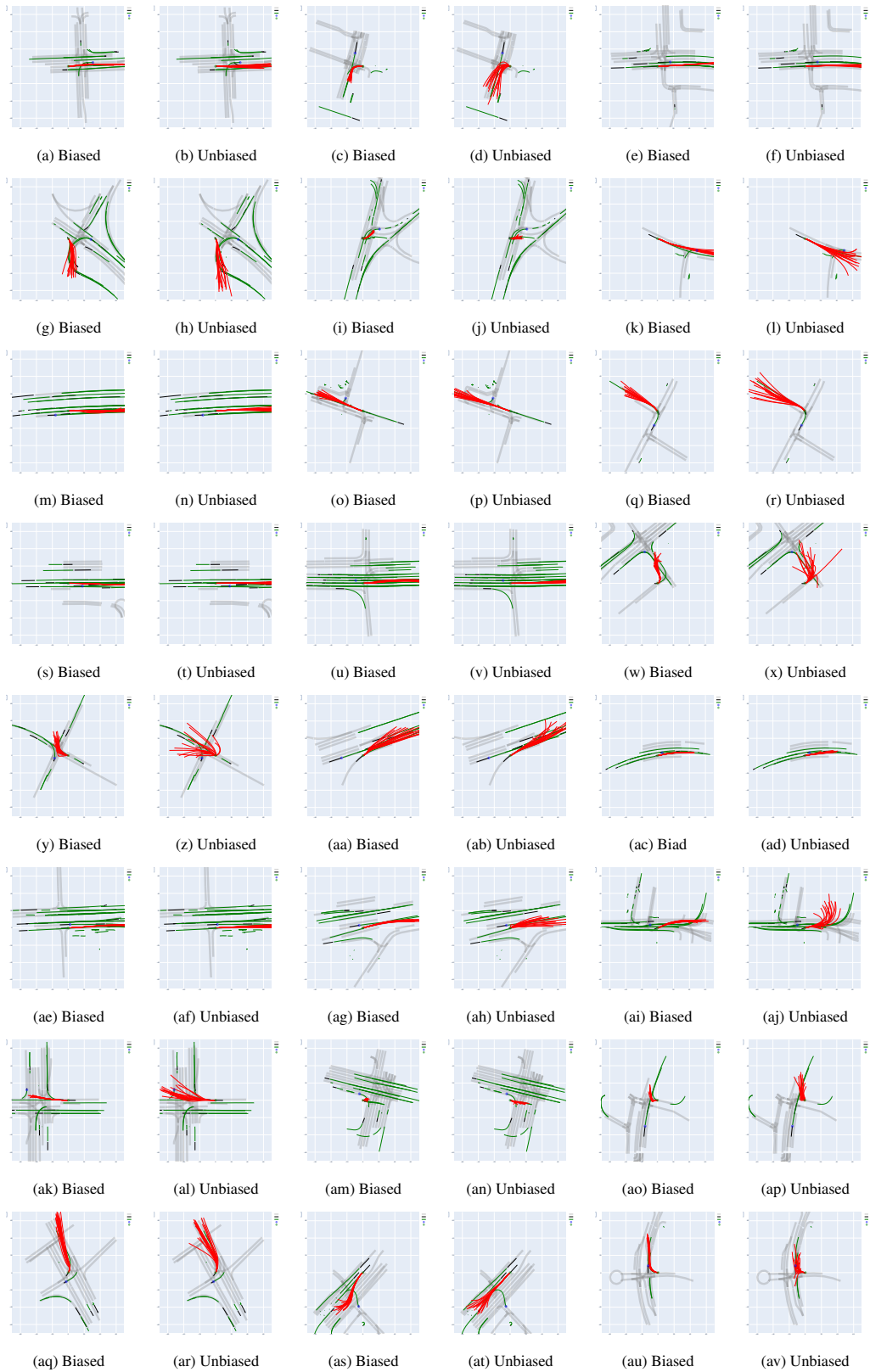}
        }
        \caption{\cap}
        \label{subfig:\i}
        \vspace{12pt}
    \end{subfigure}\quad
}
\caption{Sample results on WOMD. Each sample is depicted twice. Once with 16 samples of our biased model forecasts at risk-level $\sigma=0.95$, and once with 16 samples of our unbiased model forecasts.}
\label{fig:waymo_samples}
\end{figure}

\cref{fig:waymo_samples} depicts many samples from the Waymo open motion dataset. It shows our results with biased and unbiased forecasts.
Producing a good forecasting model is challenging, and as shown in the unbiased samples, our model counts a number of imperfections. One striking limitation is that the correlation between the map layout and the vehicle trajectories is not well captured (no lane-following behavior). This is a known limitation of trajectory forecasting models~\cite{bahari2022vehicle} that we do not solve with our proposed forecasting model.
In fact, merely defining what should be a ``good'' forecasting model is challenging. Good properties from a forecasting model could be: better interpretability (for example with a disentangled latent representation), higher dataset likelihood, more diverse forecasts, feasible predicted dynamics (acceleration and turn-rate in bounds), feasible predicted states (within drivable area bounds), accounting for many observations (agent types, different road elements, input uncertainties, etc...), accounting for unobserved hypothesis (ego plan, ``what-if''~\cite{khandelwal2020if}, given maneuver, etc...).
Improving the forecasting model before training the biasing encoder would probably improve our results in the same desired direction of improvement. In particular, restricting the state and dynamics to realistic ones might avoid some failure modes of the biased forecasts that achieve greater risk with unrealistic trajectories (see subfigures (g), (w), (y), (ao)).
Learning to bias a model with an interpretable latent space could bring some exciting results by producing interpretable directions for cost in the latent space, thus yielding interpretable reasons to be cautious.
A base forecasting model with a tighter likelihood fit would restrict the biased forecasts to ones that are closer to the trajectory distribution of the dataset, while a base model predicting more diverse trajectories would lead to a biased model that accounts for more possible costly outcomes.
Improving all these aspects with the same model is challenging, but depending on the desired usage, the focus can be put on one or another good property.

\section{Latent space exploration}
\label{sec:latent_explore}

In Section \ref{sec: experiments}, we performed didactic experiments with a two dimensional latent space. This allows us to explore the latent representation with plots in the latent space.

\begin{figure}[!h]
    \centering
    \adjustbox{width=0.5\columnwidth, trim=0 0 0 0, clip}{%
        \includegraphics{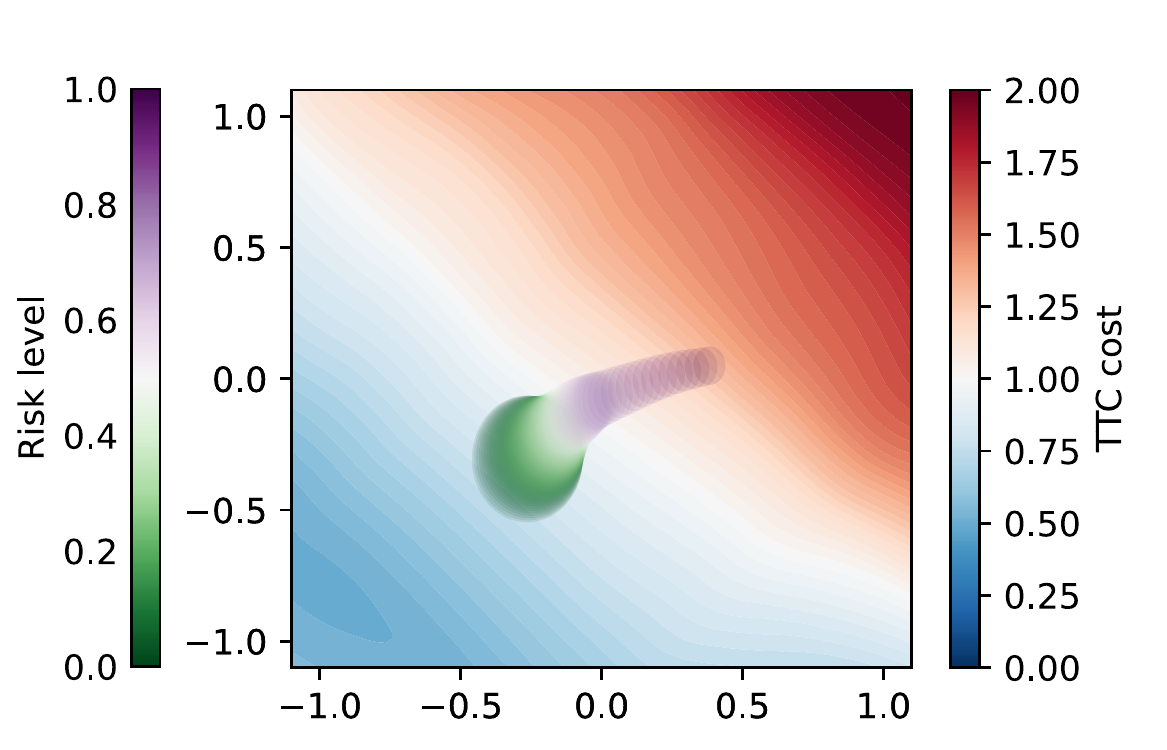}
    }
    \vspace{12pt}
    \caption{
            Representation of the latent space in the situation depicted in \cref{fig: road_scene}. On the blue to red scale, the cost of the decoded trajectories matching different latent values is mapped. On the green to purple scale, the encoded latent distributions from the biased encoder at different risk levels are represented as one-standard-deviation ellipses.
    }
    \label{fig:latent_representation}
\end{figure}

\cref{fig:latent_representation} shows the cost associated with different latent samples, and a representation of the biased distributions at different risk-levels.
It takes place in the situation described in \cref{fig: road_scene}. The biased distributions in latent space are represented with ellipses centered on the distribution mean $\mu$ with radii given by the standard deviation $\sqrt{\operatorname{diag}(\Sigma)}$ (the square root is applied on both terms). Notice that the unbiased latent distribution is neither centered on $(0, 0)$ nor has unit variance. This is because we did not train the model with respect to a prior $\mathcal{N}(0, I)$, but instead with respect to an inferred prior $\mathcal{N}(\mu|_x, \Sigma|_x)$. As the risk-level is increased, the latent mean is moved further from its starting point and the standard deviation becomes smaller. The biased distribution converges towards a riskier area in the nearby latent space. We can see on \cref{fig:latent_representation} that the direction of higher cost coincides with the direction of $\mu^{(b)}$ as the risk-level $\sigma$ is increased.

\section{Time-to-collision cost}
\label{sec: ttc_cost}
If a constant velocity model predicts an \textit{imminent} collision, we can consider that it is not avoidable and the cost of the situation is high.
 This motivates the use of a time to collision cost (TTC cost) that corrects some short-comings of the distance based cost. It considers the danger due to greater relative speed and the anisotropy due to the velocity orientation.
 
 Let us consider two agents $i$ and $j$ at positions $(x_i, y_i)$, $(x_j, y_j)$ and with velocities $(v^x_i, v^y_i)$, $(v^x_j, v^y_j)$. We want to compute a TTC cost for their relative states. The relative positions and velocities are defined as:
 \begin{align}
     (dx, dy) &\triangleq (x^i - x^j, y^i - y^j),\\
     (dv_x, dv_y) &\triangleq (v_x^i - v_x^j, v_y^i - v_y^j).
 \end{align}
 
 Under the constant velocity assumption, the evolution of relative squared distance is given by:
 \begin{equation}
     d^2(t) = (dv_x t + dx)^2 + (dv_y t + dy)^2.
 \end{equation}
 
 Finding the time to collision is finding $t_\text{col}$ at which $d^2(t_\text{col}) = 0$.
 Writing the relative speed $dv = \sqrt{dv_x^2 + dv_y^2}$ and the initial distance $d_0 = \sqrt{dx^2 + dy^2}$, we must solve the quadratic equation:
 \begin{equation}
     t_\text{col}^2 + 2t_\text{col}\frac{dv_x dx + dv_y dy}{dv^2} + \frac{d_0^2}{dv^2} = 0.
 \end{equation}
 
 The solutions in $\mathbb{C}$ are:
 \begin{align}
     t_\text{col}^{+} &= -\frac{dv_x dx + dv_y dy}{dv^2} + i\frac{dv_x dy - dv_y dx}{dv^2},\\
     t_\text{col}^{-} &= -\frac{dv_x dx + dv_y dy}{dv^2} - i\frac{dv_x dy - dv_y dx}{dv^2}.
 \end{align}
 
 Of course, not every situation leads to a collision and when there is a collision, there is only one time of occurrence. Thus, we are not surprised that there is only one real solution when there is a solution. However, this assumes a collision only when the distance between agents is exactly $0$. We should relax this assumption to consider a cost when the relative distance is low.
 
 The time of the lowest relative distance is given by the real part of the solution, and the distance at that time is given by the imaginary part multiplied by the velocity:
 \begin{align}
     t_{\tilde{\text{col}}} &= -\frac{dv_x dx + dv_y dy}{dv^2},\\
     d^2(t_{\tilde{\text{col}}}) &= \frac{(dv_x dy - dv_y dx)^2}{dv^2}.
 \end{align}
 
 Using these two values, we want to define a cost function that penalize low relative distance in a near future. However, we must first consider two problematic cases: when $dv$ is close to $0$ and when $t_{\tilde{\text{col}}}$ is negative.
 
 \textbf{When $dv$ is close to $0$,} the TTC would become large which is a low cost situation. We can simply consider a lowest possible value for $dv$ and use $\tilde{dv} \triangleq \text{max}(dv, \varepsilon)$. This overestimates the cost in very low cost situations.
 
 \textbf{When $t_{\tilde{\text{col}}}$ is negative,} the relative distance between the agents is increasing. Therefore, the actual TTC is infinite. Accounting for uncertainties, in this case, we do not set the TTC to infinity but fall back to the distance based cost by setting TTC to 0 and the distance at collision to the current distance.
 
 The values used to define the cost are now:
 
  \begin{align}
     \tilde{t}_{\tilde{\text{col}}} &\triangleq 
     \begin{cases}
        -\frac{dv_x dx + dv_y dy}{\tilde{dv}^2} & \text{if } t_{\tilde{\text{col}}} \geq 0,\\
        ~~~~~~~~0              &\text{otherwise},
    \end{cases}
 \end{align}
 
 \begin{align}
      \tilde{d}^2_{\tilde{\text{col}}} &\triangleq
      \begin{cases}
         \frac{(dv_x dy - dv_y dx)^2}{\tilde{dv}^2} & \text{if } t_{\tilde{\text{col}}} \geq 0,\\
         ~~~~dx^2 + dy^2 &\text{otherwise}.
      \end{cases}
 \end{align}
 
 Using these two values, the instantaneous cost at $t=0$ is given by :
  \begin{equation}
     J = \exp \left( -\frac{\tilde{t}_{\tilde{\text{col}}}^2}{2\lambda_t}  -\frac{\tilde{d}^2_{\tilde{\text{col}}}}{2\lambda_d} \right).
 \end{equation}
 
 For a finite-time prediction from $t=0$ to the final time-step $T$, the cost of a trajectory is the average of the instantaneous costs along the trajectory:
 \begin{equation}
     J = \frac{1}{T} \sum_{t=0}^T\exp \left( -\frac{\tilde{t}_{\tilde{\text{col}}}^2(t)}{2\lambda_t}  -\frac{\tilde{d}^2_{\tilde{\text{col}}}(t)}{2\lambda_d} \right),
 \end{equation}
 
 with a time bandwidth parameter $\lambda_t$, and a distance bandwidth parameter $\lambda_d$. This cost is high if and only if the time to collision is low compared to the time bandwidth and the distance to collision is low compared to the distance bandwidth.
 This formula uses a constant velocity assumption that only holds at short time horizon, therefore a rather small time bandwidth should be chosen.
 
 \begin{figure}[ht]
    \centering
    \adjustbox{width=\columnwidth, trim=0 0 0 0, clip}{%
        \includegraphics{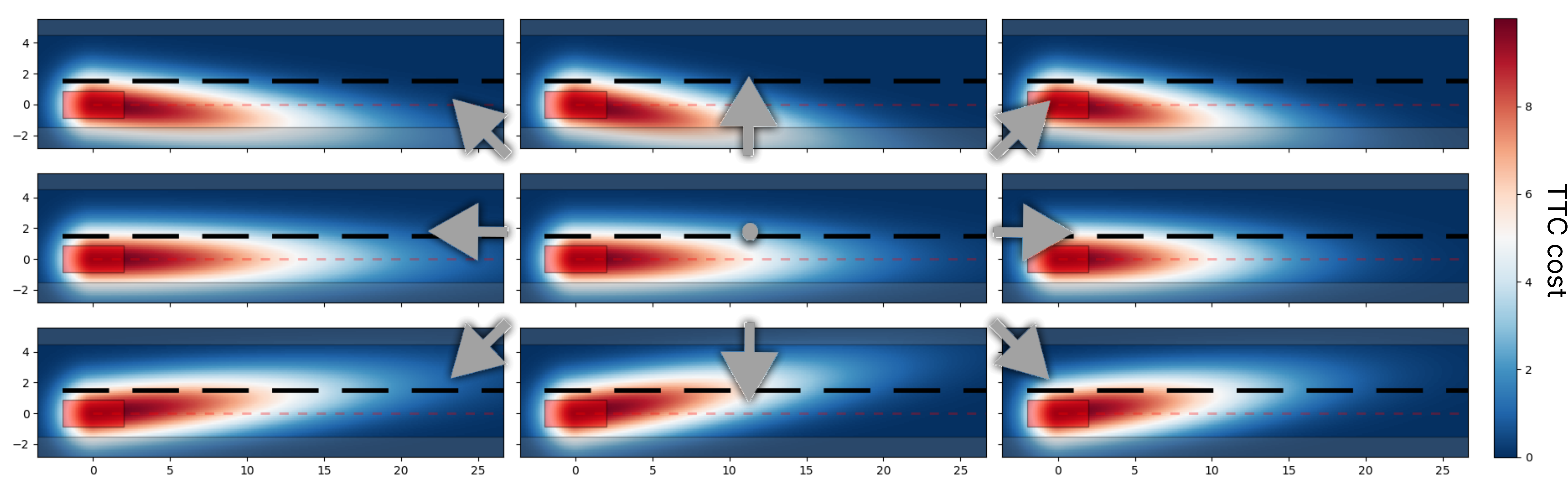}
    }
    \caption{Cost maps of a single road scene with a car going at 14m/s from left to right. The instantaneous TTC cost associated with a second agent is represented by colors from red to blue at the different positions. In each image, the cost is computed at $t=0$ for a second agent going at 2 m/s with an orientation represented by the arrows (except the center picture that considers a static second agent). The time bandwidth is 0.2 and the distance bandwidths is 2.}
    \label{fig:distance_cost}
\end{figure}

\end{document}